\documentclass[a4paper,fleqn]{cas-dc}

\usepackage[numbers]{natbib}

\usepackage{pifont}
\usepackage{array}
\usepackage{algorithm}
\usepackage{algorithmic}
\usepackage{comment}

\usepackage{placeins}

\def\tsc#1{\csdef{#1}{\textsc{\lowercase{#1}}\xspace}}
\tsc{WGM}
\tsc{QE}
\begin{document}
\let\WriteBookmarks\relax
\def\floatpagepagefraction{1}
\def\textpagefraction{.001}

\shorttitle{FairEnc}    

\shortauthors{}  

\title [mode = title]{FairEnc: A Fair Vision-Language Model with Fair Vision and Text Encoders for Glaucoma Detection}




\author[1,4]{Mohamed Elhabebe}%[<options>]

%\cormark[1]


\ead{mohamed.elhabebe@oulu.fi}


\credit{Conceptualization, Methodology, Software, Validation, Visualization, Writing -- original draft, Writing -- review \& editing}

\affiliation[1]{organization={Center for Machine Vision and Signal Analysis, University of Oulu},
           % city={},
%          citysep={}, % Uncomment if no comma needed between city and postcode
            country={Finland}}

\author[2]{Ayman El-Baz}%[]

% Footnote of the second author

% Email id of the second author
\ead{ayman.elbaz@louisville.edu}


\credit{Data Curation, Writing -- review \& editing}

\affiliation[2]{organization={Department of Bioengineering, University of Louisville},
            %addressline={}, 
            %city={},
%          citysep={}, % Uncomment if no comma needed between city and postcode
            %postcode={}, 
            %state={},
            country={USA}}

\affiliation[3]{organization={The Department of Physics and Engineering, UiT The Arctic University of Norway},
           % city={},
%          citysep={}, % Uncomment if no comma needed between city and postcode
            country={Norway}}

\affiliation[4]{organization={Computer and Systems Engineering Department, Alexandria University},
           % city={},
%          citysep={}, % Uncomment if no comma needed between city and postcode
            country={Egypt}}
            
\author[3,1]{Qing Liu}%[<options>]

\cormark[1]

%\fnmark[3]

\ead{qing.liu@oulu.fi}


\credit{Conceptualization, Methodology, Visualization, Funding acquisition, Writing -- review \& editing, Supervision}

\cortext[1]{Corresponding author}



\begin{abstract}
Automated glaucoma detection is critical for preventing irreversible vision loss and reducing the burden on healthcare systems. However, ensuring fairness across diverse patient populations remains a significant challenge. In this paper, we propose \textbf{FairEnc}, a fair pretraining method for vision‑language models (VLMs) that enables simultaneous debiasing across multiple sensitive attributes. FairEnc jointly mitigates biases in both textual and visual modalities with respect to multiple sensitive attributes, including race, gender, ethnicity, and language. Specifically, for the textual encoder, we leverage a large language model to generate synthetic clinical descriptions with varied sensitive attributes while preserving disease semantics, and employ a contrastive alignment objective to encourage demographic-invariant representations. For the visual encoder, we propose a dual-level fairness strategy that combines mutual information regularization to reduce statistical dependence between learned features and demographic groups, with multi‑discriminator adversarial debiasing. Comprehensive experiments on the publicly available Harvard-FairVLMed dataset demonstrate that FairEnc effectively reduces demographic disparity as measured by DPD and DEOdds while achieving strong diagnostic performance under both zero-shot and linear probing evaluations. Additional experiments on the private FairFundus dataset show that FairEnc consistently preserves fairness advantages under cross-domain and cross-modality settings and maintains diagnostic performance within a competitive range. These results highlight FairEnc’s ability to generalize fairness under distribution shifts, supporting its potential for more equitable deployment in real-world clinical settings. Our codebase and synthetic clinical notes are available at this \href{https://github.com/Mohamed-Elhabebe/FairEnc}{\textcolor{blue}{\underline{Link}}}.
\end{abstract}


\maketitle 

\section{Intoduction}
\label{sec:intro}

Glaucoma is one of the major reasons behind irreversible blindness globally~\cite{GBlind-PO-2015}, and there has been remarkable progress in deep learning algorithms for glaucoma diagnosis in recent years~\cite{GViT-ICPR-2022, GDPooled-IO-2026}. However, fairness concerns have gained increasing attention~\cite{HarvardGF-TMI-2024} particularly in the healthcare domain where data are often collected from high-income nations. Therefore, taking fairness into consideration becomes fundamental for the ethical deployment of AI algorithms for glaucoma diagnostic.

\begin{figure}%[]
    \centering
    \includegraphics[width=\linewidth]{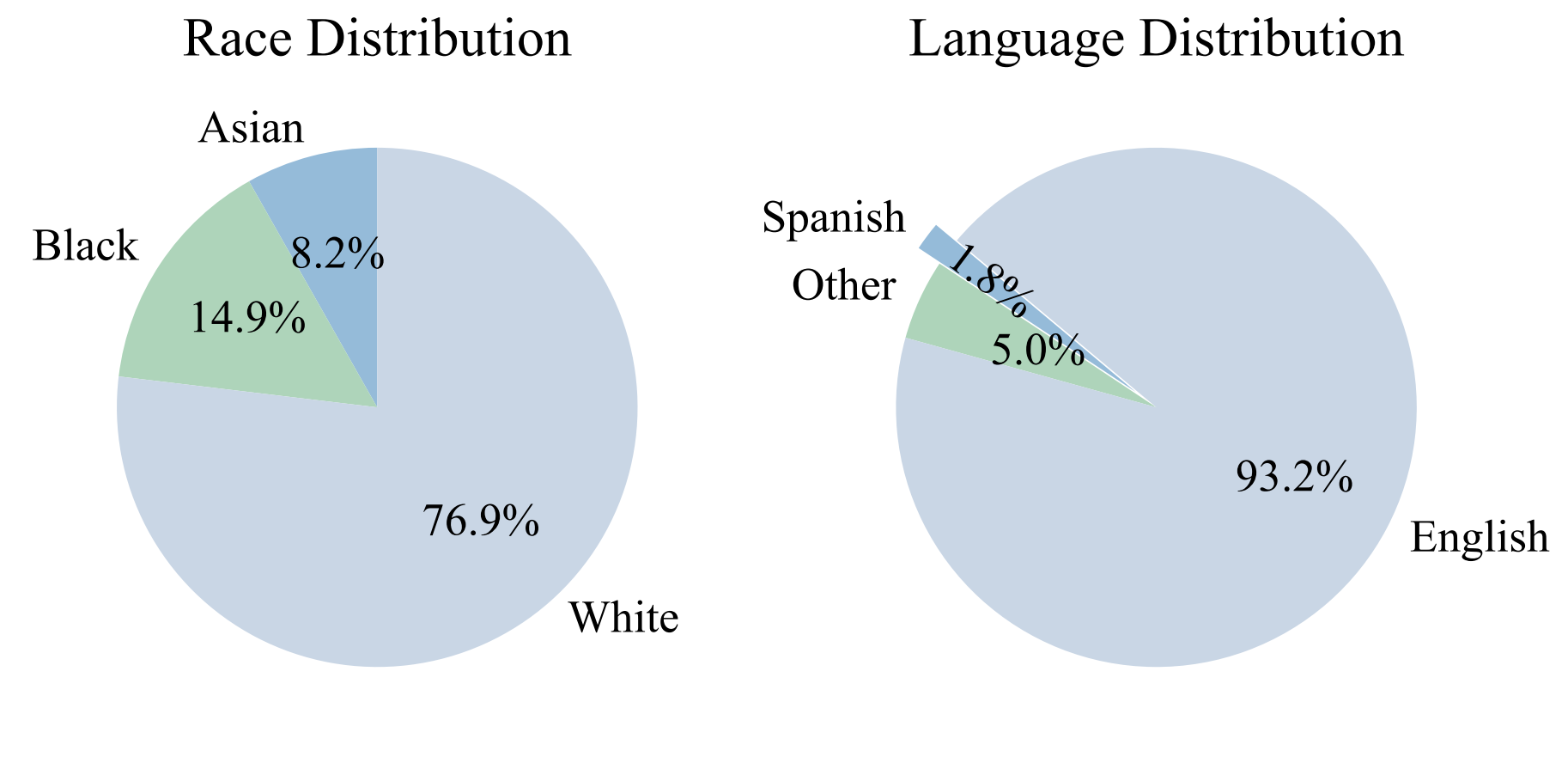}
    \caption{Race (left) and language (right) distributions of samples in Harvard-FairVLMed \cite{Fairclip-CVPR-2024}.}
    \label{fig:harvard-demographics-analysis}
\end{figure}

\begin{figure}%[]
    \centering
    \includegraphics[width=\linewidth]{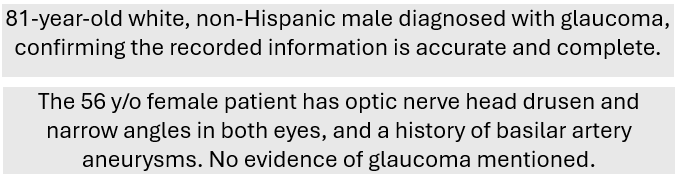}
    \caption{Examples from Harvard-FairVLMed Dataset \cite{Fairclip-CVPR-2024} for Clincal Notes that Include Demographic Information.}
    \label{fig:harvard-text-examples}
\end{figure}

% illustrating causes raising the biases 
Vision-language models (VLMs), such as CLIP~\cite{CLIP-ICML-2021} and BLIP \cite{BLIP-ICML-2022}, have recently demonstrated strong adaptability in various downstream tasks and gained a lot of attention from the medical domain where annotated data is limited and expensive to acquire~\cite{VLMs_Survey-PAMI-2024}. Specifically, VLMs offer the potential to combine clinical text (e.g., diagnostic notes) with medical imaging data through a text encoder and an image encoder respectively, introducing new fairness challenges. First, the data collected is biased. Taking Harvard-FairVLMed~\cite{Fairclip-CVPR-2024} as an example, as shown in Figure~\ref{fig:harvard-demographics-analysis}, the race, and language distributions are highly imbalanced. Training with such imbalanced data results in biased models. Second, demographic information, such as patient age, gender, and ethnicity is included in the clinical texts as shown in Figure~\ref{fig:harvard-text-examples}. Therefore, aligning visual features with these textual features makes spurious correlations with demographics more prominent.

% how do sotas solve the isses mentioned above
To mitigate the biases caused by the imbalanced data during the pretraining stage of the medical VLMs, as pioneers, FairCLIP~\cite{Fairclip-CVPR-2024} introduced the Harvard-FairVLMed dataset as the first benchmark for fair glaucoma detection, and proposed a bias mitigation technique where the feature distribution of each subgroup is enforced to align with the feature distribution of all samples within each mini-batch via minimizing the Sinkhorn distance~\cite{Sinkhorn-NeurIPS-2013}. Inspired by FairCLIP~\cite{Fairclip-CVPR-2024}, Robust FairCLIP ~\cite{RobustFairCLIP-MIPR-2025} takes the variations of sample pairs into consideration and down-weights the bad pairs while FairMOE~\cite{FairMoE-MICCAI-2025} employs multiple experts to mitigate the bias. While these methods seek fair features by pretraining separate models for each demographic attribute, they fail to achieve fairness across all attributes within a single model. This limits their applicability in real-world clinical scenarios, where equity must be achieved across all attributes. Furthermore, they overlook biases originating from clinical notes, resulting in suboptimal performance.

% how does our proposed method solve the issues mentioned above
% fisrt module in the proposed solution
In this paper, we introduce \textbf{FairEnc}, a novel fair pretraining method for vision-language models that simultaneously mitigates bias across multiple demographic attributes. To mitigate biases originating from the demographic cues in clinical notes, we propose to leverage Qwen~\cite{Qwen-arXiv-2025} to generate synthetic variants of each clinical note that either exclude demographic cues or augmented with randomized demographic cues. To enforce the text encoder focus on the clinical contents associated with diseases rather than the demographic cues, we employ a contrastive loss to align the features of these variants. 
Whereas to mitigate biases of the visual representations, we propose fair image encoder. On one hand, we learn a fair visual dictionary by minimizing mutual information (MI) between the visual feature distribution and demographic attribute distribution. On the other hand, we propose to adversely learn multiple sensitive attribute discriminators to enforce the visual encoder only learn features for disease classification rather than for demographic attribute classification. Together, these components enable FairEnc to learn disease-relevant representations while reducing spurious correlations with demographic attributes across both textual and visual modalities.

We evaluate \textbf{FairEnc} under zero-shot and linear probing configurations for glaucoma detection and compare it with state-of-the-art methods in terms of both detection performance and fairness on the publicly available dataset Harvard-FairVLMed~\cite{Fairclip-CVPR-2024}. Experimental results demonstrate that our \textbf{FairEnc} achieves better trade-off between glaucoma detection performances and fairness across multiple demographic attributes. To validate the generalization of \textbf{FairEnc}, we present a new dataset, \textbf{FairFundus}, which contains colorful retinal fundus images and clinical notes. Our main contributions are summarized as follows:
\begin{itemize}
    \item We propose \textbf{FairEnc}, a novel and intuitive fair pretraining method for medical vision-language models that achieves simultaneous debiasing across multiple demographic attributes within a single model.
    \item We propose the fair text encoder to mitigate the bias originating from clinical notes via contrasting synthetic clinical notes with random demographic attribute cues.
    \item We propose the fair vision encoder that decorrelates feature representations from demographic attributes via the proposed mutual information regularization and adversarial debiasing.
    \item We introduce the \textbf{FairFundus} dataset and provide a comprehensive evaluation demonstrating the generalization of FairEnc’s fairness advantages across dataset and imaging modality.
\end{itemize}

\section{Related Work}
\label{sec:related_work}
\subsection{Sensitive Attribute Debiasing for Fully Supervised Learning Methods}

\textbf{Sensitive Attribute Debiasing with Attribute Labels.} Regularization based methods usually aim to achieve a balance between regularization terms about fairness and performance-related loss terms during model training. For example, GroupDRO \cite{GroupDRO-ICLR-2020} adopts regularization techniques e.g. $l_2$ penalties and early stopping to improve the performances of the worst-group, thereby improving the group fairness in model generalization ~\cite{GroupDRO-ICLR-2020}. In FCRO~\cite{FCRO-IPMI-2023}, task-relevant and sensitive attribute features are enforced to be orthogonal via minimizing correlations across both samples and feature dimensions so that fair features are learned. Differently, re-weighting based methods improve the fairness via up-weighting underrepresented groups. For example, in FairBatch~\cite{Fairbatch-ICLR-2021}, a batch selection algorithm is developed to adjust the batch sizes w.r.t. sensitive groups based on fairness measures. To combine advantages of regularization-based and re-weighting based methods, FairDRO~\cite{FairDRO-ICLR-2023} is proposed to incorporate a regularization term for a well justified group fairness notion in the training objective, and to optimize the objective using a principled re-weighting learning scheme.  

\textbf{Sensitive Attribute Debiasing with Partial Attribute Labels.} To address the practical challenge that the group labels are partially annotated, Confidence-based Group Label assignment (CGL)~\cite{CGL-CVPR-2022} is proposed, which employs an auxiliary classifier to assign pseudo group labels to group-unlabeled samples. CGL ~\cite{CGL-CVPR-2022} is flexible and compatible with other fair machine learning methods, but is sensitive to the confidence threshold. To mitigate pseudo-label noise that arises when only a small portion of sensitive attributes is observed, PLEA~\cite{PLEA-ICONIP-2024} reweights samples inversely to the estimated sensitive-attribute pseudo-label error and computes a weighted fairness loss in bins, yielding notable gains in group fairness with minimal accuracy loss. Complementarily, FairCL~\cite{FairCL-ICLR-2023} tackles the partially annotated on representation-learning problem where it constructs contrastive pairs that share task-relevant content while differing along the sensitive attribute, and leverages unlabeled data to learn fairer features under limited group supervision. In application settings mixing known and unknown attributes, FairLISA~\cite{FairLISA-NeurIPS-2023} provides a general adversarial framework that couples objectives on the labeled and unlabeled subsets, enabling fair user modeling with scarce sensitive annotations.

\textbf{Sensitive Attribute Debiasing without Attribute Labels.} Sensitive attribute annotations are often expensive to obtain or even not feasible to access due to the privacy protection etc. To develop fair machine learning methods without sensitive attribute labels, Liu et al. observe that ERM models tend to misclassify worst-group samples and propose Just Train Twice (JTT)~\cite{JTT-ICML-2021}. JTT heuristically identifies worst-group samples from misclassified samples during the first training stage, then up-weight them during the second training stage. To better re-weight samples, Lahoti et al.~\cite{ARL-NeurIPS-2020} propose to adversarially assign the weights for computationally-identifiable regions of errors while Qiu et al. propose to retrain only the final classification layer of a pretrained ERM model using weights proportional to prediction errors ~\cite{AFR-ICML-2023}. To avoid extensive hyperparameter search, Zarlenga et al. propose Targeted Augmentations for Bias Mitigation (TAB)~\cite{TAB-ECCV-2024} which generates a group balanced training set with a helper model to train a fair model. Whereas, Kim et al. propose to employ a committee of classifiers to identify and weight bias-conflicting samples ~\cite{biased-committee-NeurIPS-2022}. 

\subsection{Fair Vision-Language Model Pretraining in Medical Domain}
Fair pretraining of medical VLMs is essential due to its direct impact on patient outcomes. As pioneers, FairCLIP~\cite{Fairclip-CVPR-2024} mitigates biases during pretraining stage and adapts the CLIP model with biased retinal images and clinical notes fairly via minimizing the distance between the feature distribution of samples within each batch and that of the samples belonging to each specific subgroup. To take the variations of image-text pairs in consideration during pretraining, Robust FairCLIP~\cite{RobustFairCLIP-MIPR-2025} is proposed which dynamically mines bad pairs in each epoch and down-weights noisy or faulty image-text pairs, improving robustness to GPT-generated textual noise. FairMoE~\cite{FairMoE-MICCAI-2025} further extends FairCLIP~\cite{Fairclip-CVPR-2024} by incorporating fairness-oriented mixture of experts into both visual and text encoders, along with a fairness-oriented loss to align expert gate weight dispersions across subgroups. While FairMoE~\cite{FairMoE-MICCAI-2025} improves learning capacity and debiases both modalities, it does not guarantee fairness improvements, and retains the group homogeneity assumption.

\section{Methodology}

\subsection{Problem Formulation and Framework Overview}
\label{subsec:problem-formulation}

Suppose that we are given the VLM that incorporates a visual encoder $\phi_{img}$ and a text encoder $\phi_{txt}$ and pretrained with natural scene images and text descriptions, a set of $M$ sensitive attributes $\left\{A_{m}\right\}_{m = 1}^M$, and the training set $\left\{\left(\mathbf{X}_{img}^{(i)}, \mathbf{X}_{txt}^{(i)}, \left\{A_{m}^{(i)}\right\}_{m = 1}^M\right) \right\}_{i=1}^N$ comprising $N$ samples, where $i$ represents the sample index, $\mathbf{X}_{img}^{(i)}$ is the retinal image belonging to sample $i$, $\mathbf{X}_{txt}^{(i)}$ is the corresponding text clinical note, and $A_m^{(i)}\in \mathcal{A}_m$ is the group value of sample $i$ corresponding to the attribute $A_m$ with attribute space $\mathcal{A}_m$. Our goal is to continually pretrain the VLM fairly using the training set so that the VLM is able to learn fair features over sensitive attributes for glaucoma detection.

\begin{figure*}[!t]
    \centering
    \includegraphics[width=0.8\linewidth]{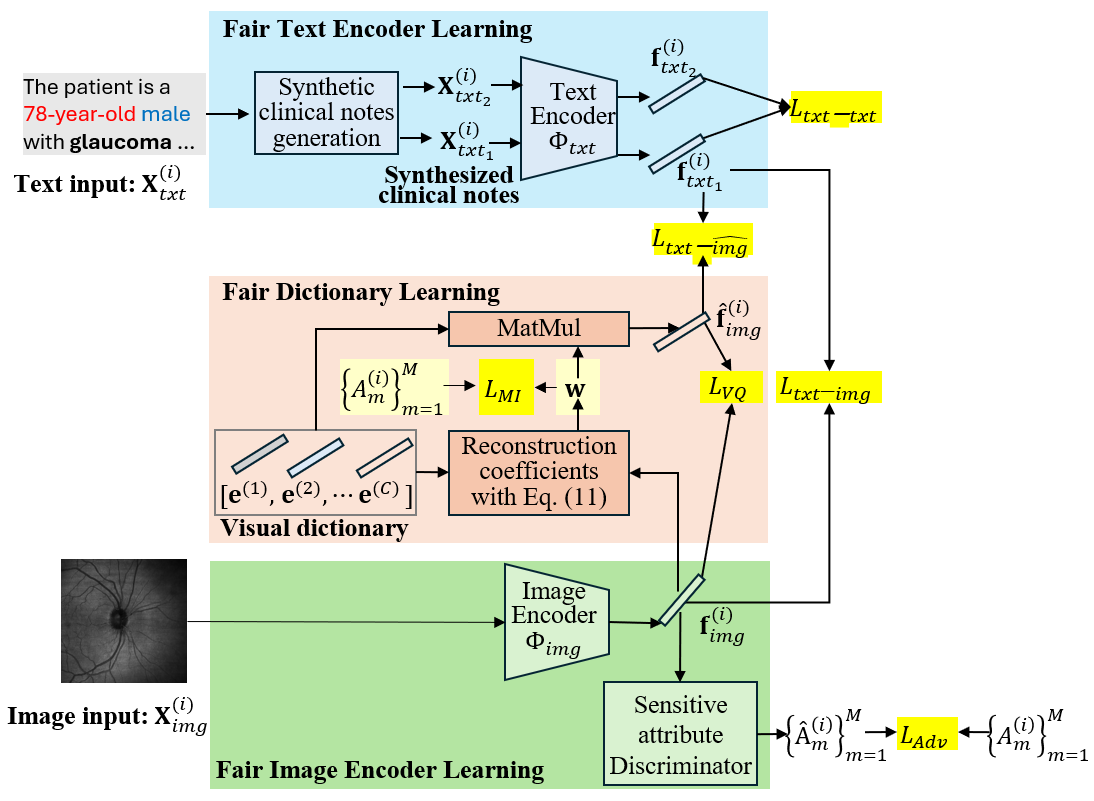}
    \caption{FairEnc Overall Architecture}
    \label{fig:overall-arch}
\end{figure*}

An overview of our complete framework is illustrated in Figure~\ref{fig:overall-arch}. The process begins with the independent generation of synthetic clinical notes using Qwen~\cite{Qwen-arXiv-2025}. Each input sample to the VLM consists of a retinal image paired with two corresponding versions of synthetic clinical notes. Contrastive learning is employed to debias text representations whereas both adversarial learning and mutual information minimization are employed to debias visual representations.

\subsection{Fair Text Encoder}
\label{subsec:fair-text}

In medical VLMs, clinical notes associated with the medical images are fed into the text encoder and generates corresponding text representations. Then these text representations are aligned with the visual representations in a  latent space. Typically, the demographic cues in the clinical notes are also encoded. Aligning the visual features with the text features compromises fairness in vision-based downstream tasks. To mitigate the influence of demographic information in clinical notes, we employ a LLM to generate multiple textual variants for each sample. These synthetic clinical notes are created either by removing demographic information or by substituting it with randomly sampled demographic attributes. We then apply contrastive learning to align the representations of these variants, thereby encouraging the text encoder to focus on intrinsic disease-related semantics rather than learning spurious correlations associated with demographic cues.

\textbf{Synthetic clinical notes generation.} We leverage the capabilities of Qwen~\cite{Qwen-arXiv-2025} to generate two kinds of clinical notes: neutral clinical notes and synthetic clinical notes with random demographic information. For the clinical note $\mathbf{X}_{txt}^{(i)}$ from sample $i$, we carefully design a prompt $Prompt_{Neutr}$ and feed it to Qwen~\cite{Qwen-arXiv-2025} to remove all demographic information from the original clinical note:
\begin{equation}
    \mathbf{X}_{Neutr\text{-}txt}^{(i)} = \text{Qwen}\left(\mathbf{X}_{txt}^{(i)}, Prompt_{Neutr}\right)\;.
\end{equation}
In our experiments, we set $Prompt_{Neutr}$ to \texttt{\lq\lq Rewrite the following clinical note to exclude any details related to the patient's race, gender, ethnicity, preferred or native language, or age:\rq\rq}.

Subsequently, with the synthetic clinical notes without any information about demographic attributes, we prompt Qwen~\cite{Qwen-arXiv-2025} with  $Prompt_{random}$ to generate $K$ clinical notes with distinct random combination of demographic attributes, including values for race, gender, ethnicity, preferred language, or age: 
\begin{equation}
    \left\{\mathbf{X}_{Syn\text{-}txt,k}^{(i)}\right\}_{k=1}^K = \text{Qwen}\left(\mathbf{X}_{Neutr\text{-}txt}^{(i)}, Prompt_{random}\right)\;.
\end{equation}
To maintain the consistent semantics about the diseases and manifestations, we insert random demographic attributes and values, ensuring natural and fluent sentence construction. In our experiments, $K$ is set to 5 and $Prompt_{random}$ is set to \texttt{\lq\lq Rewrite the following clinical note 5 times to integrate random information about the patient's race, gender, ethnicity, preferred or native language, or age at appropriate positions in the note. You can at most add or modify two sentences. The race options are exclusively black, white, or asian. The ethnicity options are exclusively hispanic or non-hispanic. The gender options are exclusively male or female. The preferred or native language options are mainly English or Spanish and may be any spoken language. The age range is from 20 to 80. You need to randomly add information about one or more of these demographics to each version and output 5 concise sentences without any added special marks. Add a new line between each 2 consecutive versions to split them:\rq\rq}. Table~\ref{tab:text-examples} presents examples of original clinical notes from the Harvard-FairVLMed dataset, alongside their corresponding synthetic versions, both neutralized and augmented with random demographic attributes.

%The complete generation procedure is detailed in Algorithm~\ref{alg:synthetic-generation}. Despite these modifications, it is observable that the synthetic variants maintained consistent disease semantics, and demographic insertions or removals were performed with high linguistic fidelity, ensuring natural and fluent sentence construction. Table~\ref{tab:text-examples} presents examples of original clinical text descriptions from the Harvard-FairVLMed dataset, alongside their corresponding synthetic versions—both neutralized and augmented with random demographic attributes.

%During training, for each image-text pair, we randomly selected either the neutral version (with 50\% probability) or one of the five demographic variants (also with 50\% probability) as the associated textual input denoted as  \(  X_{txt1}^{(i)} \), where \( i \) represents the sample index in the batch. This strategy served a dual purpose: presenting the model with demographic-free descriptions reinforced its focus on disease-relevant features, while the randomized demographic variants conditioned the model to disregard sensitive attribute values as predictive cues. Collectively, this data augmentation approach contributed to debiasing the text encoder and mitigating the propagation of demographic bias into the visual feature representations.

\begin{table*}[h]
\centering
\caption{\textbf{Examples of generated clinical notes.} The first column presents the original text samples from the dataset. The second column contains the corresponding neutral synthetic versions. The third column displays one out of five synthetic variants, each augmented with randomly assigned demographic attributes.}
\label{tab:text-examples}
\footnotesize
\resizebox{\textwidth}{!}{
\begin{tabular}{
>{\centering\arraybackslash}p{0.3\textwidth}
>{\centering\arraybackslash}p{0.345\textwidth}
>{\centering\arraybackslash}p{0.345\textwidth}
}
\hline
\textbf{Original} & \textbf{Neutral} & \textbf{Random}\\
\hline
The patient is a 28-year-old white, non-Hispanic female. She has not been diagnosed with glaucoma. &  
The patient has not been diagnosed with glaucoma. &  
The patient has not been diagnosed with glaucoma. She is 58 years old, non-Hispanic, and speaks Spanish as her preferred language. \\
\hline
The patient shows open angle glaucoma risk in both eyes, especially with ocular hypertension. Abnormal visual field detected in left eye. Past retinal procedures were mentioned. Current plan is to begin latanoprost medication for glaucoma. Follow-up in 1 month. &  
The patient shows open angle glaucoma risk in both eyes, particularly with ocular hypertension. An abnormal visual field was detected in the left eye. Previous retinal procedures were noted. The current plan is to start latanoprost medication for glaucoma. A follow-up appointment is scheduled in one month. &  
The patient shows open angle glaucoma risk in both eyes, particularly with ocular hypertension. She is Black and is 65 years old. An abnormal visual field was detected in the left eye. Previous retinal procedures were noted. The current plan is to start latanoprost medication for glaucoma. A follow-up appointment is scheduled in one month. \\
\hline
\end{tabular}
}
\end{table*}

\textbf{Fair text encoder via contrastive learning.} To encourage the text encoder of the VLM to focus more effectively on disease-related semantics and contextual information during feature extraction while disregard the demographic information, we propose to learn the fair text embeddings via contrastive learning. 

More specifically, we first randomly select two synthesized clinical notes $ \mathbf{X}_{txt_1}^{(i)} $ and $ \mathbf{X}_{txt_2}^{(i)} $ according to the distributions defined below:
\begin{equation}
\label{eq:text-select-dist}
\begin{aligned}
\pi_1\big(\mathbf{X}_{txt_1}^{(i)} = x\big) =&
\begin{cases} 
p_{txt_1}, & \text{if } x = \mathbf{X}_{Neutr\text{-}txt}^{(i)},\\[1mm]
\dfrac{1 - p_{txt_1}}{K}, & \text{if } x = \mathbf{X}_{Syn\text{-}txt,k}^{(i)},\ k = 1, \dots, K  \;,
\end{cases} \\[2mm]
%\mathbf{X}_{txt_2}^{(i)} &\sim \mathrm{Uniform}\Big(\big\{\mathbf{X}_{Syn\text{-}txt,k}^{(i)}\big\}_{k=1}^K \setminus \big\{\mathbf{X}_{txt_1}^{(i)}\big\}\Big)
\mathbf{X}_{txt_2}^{(i)} \sim \mathrm{Uniform}&\Big(\big\{\big\{\mathbf{X}_{Neutr\text{-}txt}^{(i)}\big\} \cup \big\{\mathbf{X}_{Syn\text{-}txt,k}^{(i)}\big\}_{k=1}^K\big\} \setminus \big\{\mathbf{X}_{txt_1}^{(i)}\big\}\Big)\;.
\end{aligned}
\end{equation}
where $\pi_1$ is the distribution for the random selection of $ \mathbf{X}_{txt_1}^{(i)} $ and $p_{txt_1}$ is the predefined probability to select the synthetic clinical notes without any demographic information $\mathbf{X}_{Neutr\text{-}txt}^{(i)}$. We then feed the selected synthetic clinical notes to the text encoder and obtain the text embeddings: 
\begin{equation}
\begin{aligned}
\mathbf{f}_{txt_1}^{(i)} &= \phi_{txt}\left(\mathbf{X}_{txt_1}^{(i)}\right) \;,\\
\mathbf{f}_{txt_2}^{(i)} &= \phi_{txt}\left(\mathbf{X}_{txt_2}^{(i)}\right)\;.
\end{aligned}
\end{equation}

To enforce the text encoder to only learn semantically meaningful representations aligned with disease-related content while encouraging it disregard demographic information, we apply a contrastive loss to push away the embedding distance of synthetic clinical notes for different patients and pull close the embedding distance of synthetic clinical notes for the same patients. We adopt the NT-Xent contrastive loss \cite{SimCLR-ICML-2020}, which is computed as:
\begin{equation}
\begin{aligned}
L_{txt\text{-}txt} = &\text{NT-Xent}\left(\mathcal{S}_{+}, \mathcal{S}\right)\\
=&-\frac{1}{2B}
\sum_{(\mathbf{f},\mathbf{f}^{+})\in\mathcal{S}_+}
\log
\frac{
\exp{\left( \text{cos}(\mathbf{f},\mathbf{f}^{+}) / \tau \right)}
}{
\sum\limits_{(\mathbf{f},\mathbf{f}^{'})\in\mathcal{S}^{(\mathbf{f})}}
\exp{\left( \text{cos}(\mathbf{f},\mathbf{f}^{'}) / \tau \right)}
} \;,
\end{aligned}    
\end{equation}
where the cosine similarity measures the representation distance and $\mathcal{S}_+$ is the set of positive pairs and $\mathcal{S}$ is the set of all pairs combinations. $\mathcal{S}_+$ includes matched clinical notes of the same sample, while \( \mathcal{S} \) includes all combinations of clinical note pairs throughout the batch. We suppose the batch size is $B$, and each of the text representations $\{\mathbf{f}_{txt_1}^{(i)}, \mathbf{f}_{txt_2}^{(i)}\}_{i=1}^{B}$ is used once as an anchor and has exactly one positive counterpart. Then the sets are formally defined as:
\begin{equation}
\mathcal{S}_{+}
=
\Bigl\{
(\mathbf{f}_{txt_1}^{(i)}, \mathbf{f}_{txt_2}^{(i)}),
(\mathbf{f}_{txt_2}^{(i)}, \mathbf{f}_{txt_1}^{(i)})
\ \Big|\ i = 1,\dots,B
\Bigr\}\;,
\end{equation}
and
\begin{equation}
\mathcal{S} =
\Bigl\{ \mathcal{S}^{(\mathbf{f})} \;\Big|\; 
 \ \mathbf{f} \in \{\mathbf{f}_{txt_1}^{(i)}, \mathbf{f}_{txt_2}^{(i)}\}_{i=1}^{B} \Bigr\}\;,
\end{equation}
\begin{equation}
\mathcal{S}^{(\mathbf{f})}
=
\Bigl\{
(\mathbf{f},\mathbf{f}^{'})
\ \Big|\ \
\mathbf{f}^{'} \in 
\big\{\{\mathbf{f}_{txt_1}^{(i)},\mathbf{f}_{txt_2}^{(i)}\}_{i=1}^{B} \setminus \{\mathbf{f}\} \big\}
\Bigr\}\;.
\end{equation}

\subsection{Fair Visual Encoder}
\label{subsec:visual-debiasing}

The objective of fair training is to develop a model whose outputs are invariant to sensitive attributes. In the context of pretraining a vision-language (VL) model, the retinal image $ \mathbf{X}_{img}^{(i)} $ is fed through the visual encoder $ \Phi_{img} $ to generate its corresponding feature representations as:
\begin{equation}
     \mathbf{f}_{img}^{(i)} = \phi_{img}\left(\mathbf{X}_{img}^{(i)}\right)\;.
\end{equation}
These features are subsequently used for various vision-based downstream tasks, including classification. Consequently, the primary goal of fair fine-tuning becomes ensuring that the visual features are invariant with respect to the sensitive attributes. To achieve this, we adopt a dual-strategy approach that combines adversarial debiasing and mutual information regularization.

\subsubsection{Feature-Sensitive Attribute Decorrelation via Fair Visual Dictionary}

Mutual Information (MI) is usually used to measure the correlation between two statistical distributions. To decorrelate the visual features and the demographic attributes, we propose to minimize the MI between their  distributions. However, as the visual feature is a multivariate continuous random variable, it is highly challenging to estimate its distribution. To address this, we propose to learn a dictionary for the visual representations, then use the normalized reconstruction weights with the dictionary as a proxy distribution of the visual representation.

\textbf{Visual Dictionary Learning.} Inspired by~\cite{VQ-NeurIPS-2017}, we learn a dictionary comprising $C$ element vectors, denoted as $[\mathbf{e}^{(1)}, \cdots, \mathbf{e}^{(C)}]$, each matching the dimensionality of the visual feature vectors. Here, \( C \) is a hyperparameter representing the dictionary size. Then, we reconstruct the visual representation $\mathbf{f}_{img}^{(i)}$ via:
\begin{equation}
    \label{eq:soft-quantization}
    \hat{\mathbf{f}}_{img}^{(i)} = \sum_{c= 1}^{C}  w^{i,c} \cdot \mathbf{e}^{(c)}\;,
\end{equation}
where $w^{i,1}, \cdots, w^{i,C}$ are the reconstruction coefficients and we use the softmax normalized squared $L_2$ norm distance to calculate them which can be expressed as:
\begin{equation}  % reconstruction weight
    w^{i,c} = \frac{\exp\left(-\|\mathbf{f}_{img}^{(i)}-\mathbf{e}^{(c)}\|_{2}^{2}\right)}{\sum_{j=1}^C\exp\left(-\|\mathbf{f}_{img}^{(i)}-\mathbf{e}^{(j)}\|_{2}^{2}\right)}\;.
\end{equation}
Similar to~\cite{VQ-NeurIPS-2017}, to learn the dictionary, we minimize the discrepancy between the original and reconstructed representations via:
{\small
\begin{equation}
    \label{eq:vq-loss}
    L_{VQ} =\sum_{i=1}^B \frac{1}{B}\left(\left\|\hat{\mathbf{f}}_{img}^{(i)} - sg\left(\mathbf{f}_{img}^{(i)}\right)\right\|_{2}^{2} + \lambda_{Cmt}\left\|\mathbf{f}_{img}^{(i)} - sg\left(\hat{\mathbf{f}}_{img}^{(i)}\right)\right\|_{2}^{2}\right)\;, 
\end{equation}
}
where \( sg(\cdot) \) denotes the stop-gradient operation, preventing gradient flow during backpropagation.

\textbf{Fair Visual Dictionary Learning.} To decorrelate the visual representations and the sensitive attributes, we propose to learn a fair visual dictionary, in which the proportion distribution of element vectors is independent of the distributions of sensitive attributes. To this end, we minimize the mutual information: 
\begin{equation}
\label{eq:MI_loss}
\begin{aligned}
    L_{MI} &= \sum_{m=1}^M MI\left(\mathbf{f}_{img}; A_m\right)\\
    & = \sum_{m=1}^M\left(H\left(\mathbf{f}_{img}\right) - H\left(\mathbf{f}_{img}| A_m\right)\right)\;.
\end{aligned}    
\end{equation}
In Eq. (\ref{eq:MI_loss}), $H\left(\mathbf{f}_{img}\right)$ is the entropy of the visual features and we estimate it via:
\begin{equation}
        H\left(\mathbf{f}_{img}\right) = -\sum_{c=1}^C p\left(\mathbf{f}_{img}=\mathbf{e}^{(c)}\right)\cdot \log p\left(\mathbf{f}_{img}=\mathbf{e}^{(c)}\right)\;,
\end{equation}
where $p_{\mathbf{e}^{(c)}}$ is estimated over samples within each batch of size $B$ via: 
\begin{equation}
    \label{eq:features-dist}
    p\left(\mathbf{f}_{img}=\mathbf{e}^{(c)}\right) = \frac{1}{B} \sum_{i = 1}^{B}  w\left(\mathbf{f}_{img}^{(i)}, \mathbf{e}^{(c)}\right)\;.
\end{equation}
$H\left(\mathbf{f}_{img}|A_m\right)$ is the conditional entropy, which is computed via:
\begin{equation}
\begin{aligned}
    H\left(\mathbf{f}_{img}|A_m\right) 
    = -\sum_{c=1}^C\sum_{a_m\in \mathcal{A}_m} p\left(\mathbf{f}_{img}=\mathbf{e}^{(c)}, A_m=a_m\right)\cdot & \\
    \log p\left(\mathbf{f}_{img}=\mathbf{e}^{(c)}| A_m=a_m\right)& \\
    =-\sum_{c=1}^C\sum_{a_m\in \mathcal{A}_m} p\left(\mathbf{f}_{img}=\mathbf{e}^{(c)}| A_m=a_m\right)\cdot & p\left( A_m=a_m\right)\cdot\\
    \log p\left(\mathbf{f}_{img}=\mathbf{e}^{(c)}| A_m=a_m\right)&\;,
\end{aligned}
\end{equation}
where the marginal distribution of $A_m$ is estimated over the samples within each batch:
\begin{equation}
    p\left( A_m=a_m\right) =  \sum_{i=1}^B\frac{1}{B}\cdot \mathbb{1}_{a_m}\left(A_m^{(i)}\right)\;,
\end{equation}
and the conditional distribution is estimated via:
%\begin{equation}
%    p\left(\mathbf{f}_{img}=\mathbf{e}^{(c)}| A_m=a_m\right) =  \sum_{i=1}^B\frac{1}{B}\cdot \mathbb{1}_{a_m}\left(A_m^{(i)}\right) \cdot  w\left(\mathbf{f}_{img}^{(i)}, \mathbf{e}^{(c)}\right)\;.
%\end{equation}
\begin{equation}
    p\left(\mathbf{f}_{img}=\mathbf{e}^{(c)}| A_m=a_m\right) =  \frac{\sum_{i=1}^B \mathbb{1}_{a_m}\left(A_m^{(i)}\right) \cdot  w\left(\mathbf{f}_{img}^{(i)}, \mathbf{e}^{(c)}\right)}{\sum_{i=1}^B \mathbb{1}_{a_m}\left(A_m^{(i)}\right)}\;.
\end{equation}

\subsubsection{Feature-Sensitive Attribute Decorrelation via Adversarial Debiasing}

To encourage the visual encoder to learn features that are decorrelated from sensitive attributes, inspired by the bias mitigation framework introduced in \cite{AdvMitigating-AAAI-2018}, we propose to adversely learn multiple sensitive attribute discriminators. In detail, we feed the visual feature $\mathbf{f}_{img}^{(i)}$ to a multi-layer perceptron (MLP) with $M$ linear projection heads and get predictions for the sensitive attributes. Formally, the prediction for the $m$-$th$ attribute can be obtained via :
\begin{equation}
    \label{eq:att-pred}p\left({\hat{A}}_m^{(i)}=a_m|\mathbf{f}_{img}^{(i)}\right) = \text{softmax}\left(\text{LP}_m\left(\text{MLP}\left(\mathbf{f}_{img}^{(i)}\right)\right)\right)\;,
\end{equation}
where $\text{MLP}(\cdot)$ is the shared MLP and $\text{LP}_m$ is the linear projection for the $m$-$th$ attribute classification.

Our purpose is to learn visual representations which confuse sensitive attribute discriminators most. Thus, we minimize the sensitive attribute classification loss below: 
\begin{equation}
    \label{per-attr-sens-loss2}
    L_{att\text{-}cls} = -\frac{1}{B}  \sum_{i = 1}^{B} \sum_{m=1}^M\sum_{a_m\in \mathcal{A}_m}\mathbb{1}_{a_m}\left({A}_m^{(i)}\right) \log p\left({\hat{A}}_m^{(i)}=a_m|\mathbf{f}_{img}^{(i)}\right)\;,
\end{equation}
and consequently minimize the adversarial loss below:
\begin{equation}
    \label{per-attr-adv-loss2}
    L_{adv} = -L_{att-cls}\;.
\end{equation}

\subsection{Total Loss}
\label{subsec:total-loss}

Similar to CLIP~\cite{CLIP-ICML-2021}, we align the text representations and visual representations as well as the text representations and the reconstructed visual representations via contrastive learning. The alignment loss can be expressed as: 
\begin{equation}
    \label{align-loss}
    L_{align} = L_{txt\text{-}img} + \lambda_{txt\text{-}\hat{img}} \cdot L_{txt\text{-}\hat{img}} \;,
\end{equation}
 where $\lambda_{txt\text{-}\hat{img}}$ is the hyperparamter controlling the contribution of $L_{txt\text{-}\hat{img}}$, and
 \begin{equation}
 \label{eq:contrst-VL}
 \begin{aligned}
     L_{txt\text{-}img} =& -\sum_{i=1}^B \frac{1}{2B}\log \frac{\exp{\left(\text{cos}(\mathbf{f}^{(i)}_{txt_1}, \mathbf{f}^{(i)}_{img})/\tau\right)}}{\sum_{j=1}^B\exp{\left(\text{cos}(\mathbf{f}^{(i)}_{txt_1}, \mathbf{f}^{(j)}_{img})/\tau\right)}}\\
     & - \sum_{i=1}^B \frac{1}{2B}\log \frac{\exp{\left(\text{cos}(\mathbf{f}^{(i)}_{txt_1}, \mathbf{f}^{(i)}_{img})/\tau\right)}}{\sum_{j=1}^B\exp{\left(\text{cos}(\mathbf{f}^{(j)}_{txt_1}, \mathbf{f}^{(i)}_{img})/\tau\right)}}\;,
 \end{aligned}
 \end{equation}
and the contrastive loss between text representations and the the reconstructed visual representations is calculated in the same way as in Eq. (\ref{eq:contrst-VL}).

In summary, the total loss of our method is: 
\begin{equation}
    L_{total} = L_{align} + \lambda_{txt\text{-}txt}\cdot L_{txt\text{-}txt} + L_{VQ} + \lambda_{MI}\cdot L_{MI} + \lambda_{adv}\cdot L_{adv}\;,
\end{equation}
where $\lambda_{txt\text{-}txt}$, $\lambda_{MI}$ and $\lambda_{adv}$ are hyperparameters controlling the contributions of $L_{txt\text{-}txt}$, $L_{MI}$ and $L_{adv}$.

\section{Experiments}
We conduct experiments on two datasets: Harvard-FairVLMed ~\cite{Fairclip-CVPR-2024} which is publicly available, and FairFundus which is newly collected from Egypt.

\subsection{Datasets and Evaluation Metrics}
\textbf{Harvard-FairVLMed~\cite{Fairclip-CVPR-2024}.} This dataset comprises 10,000 samples, each consisting of scanning laser ophthalmoscopy (SLO) fundus images paired with corresponding clinical notes summarizing glaucoma diagnoses. In addition to glaucoma labels, the dataset includes demographic identity group information, namely race, gender, ethnicity, and language, which we utilize in our experiments as target debiased sensitive attributes.

\textbf{FairFundus.} To evaluate the cross-domain generalization capability of the pretrained foundation model, we introduce \textit{FairFundus}, a dataset containing 1{,}158 paired retinal color fundus photographs (CFP) and clinical notes collected from 656 patients in Egypt. All CFP images were acquired using a swept-source OCT device (Topcon DRI Triton, Topcon Corporation, Tokyo, Japan). The accompanying clinical notes describe the glaucoma status of each eye, and we provide labels for glaucoma diagnosis as well as demographic attributes including gender and age.

\begin{figure}%[]
    \centering
    \includegraphics[width=0.5\linewidth]{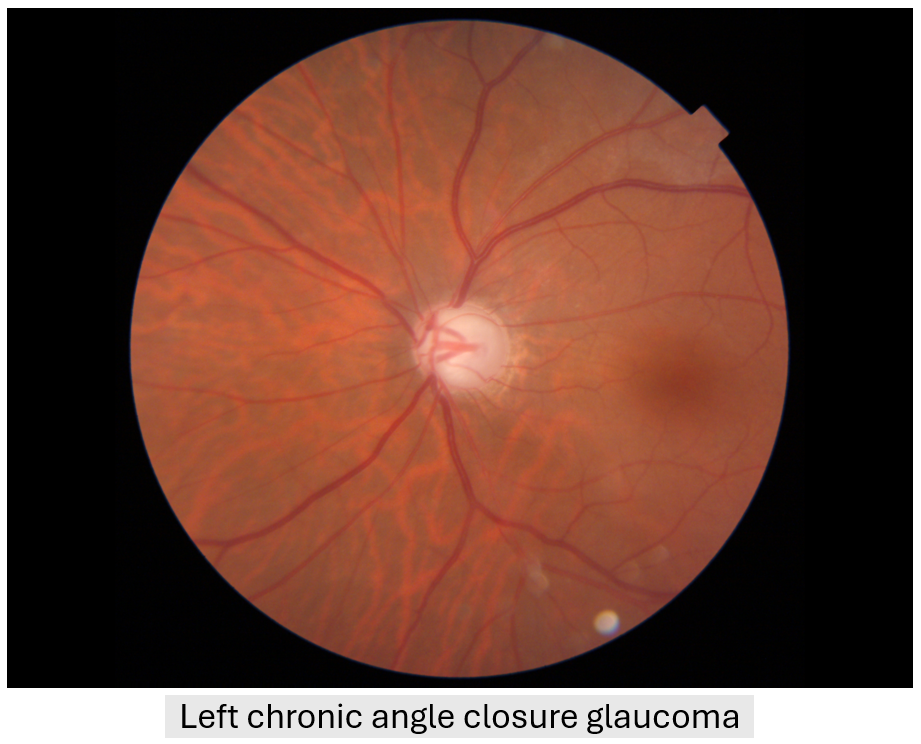}
    \caption{Example from the FairFundus dataset showing a fundus image and its corresponding clinical note.}
    \label{fig:new-dataset-example}
\end{figure}

\begin{figure}%[]
    \centering
    \includegraphics[width=\linewidth]{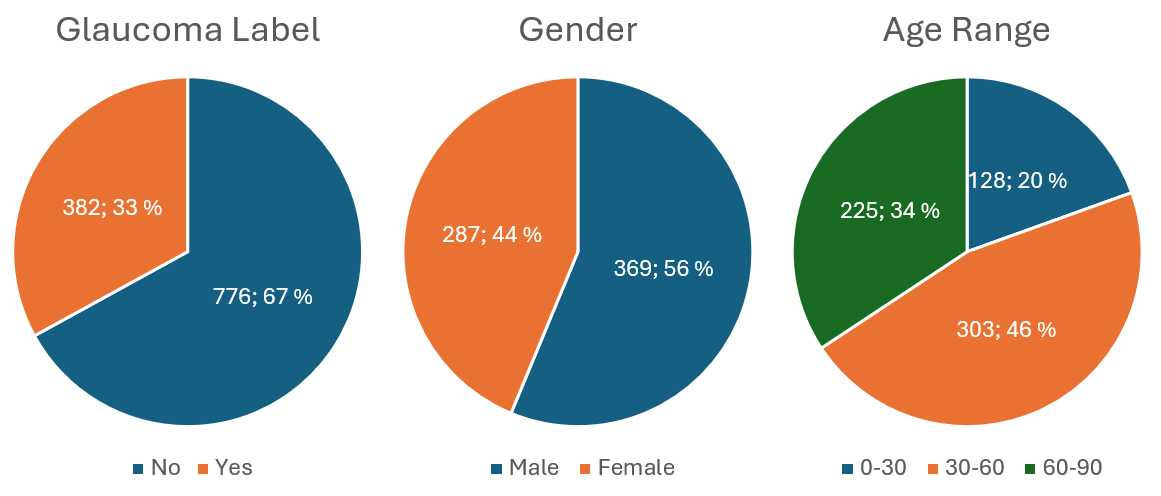}
    \caption{\textbf{Analysis of the FairFundus Dataset.} Glaucoma labels are assessed at the image level, while demographic attributes, specifically gender and age range, are analyzed at the patient level.}
    \label{fig:new-dataset-analysis}
\end{figure}

Since all data originate from a single geographic region (Egypt), the dataset exhibits limited demographic variability, with participants sharing a largely consistent ethnic background and native language  (Egyptian Arabic). As a result, attributes such as race, ethnicity, and language are not meaningfully variable in this cohort and are therefore not used as sensitive attributes in fairness evaluation. Despite this limitation, \textit{FairFundus} provides a complementary evaluation setting to Harvard-FairVLMed~\cite{Fairclip-CVPR-2024} by introducing a distinct imaging modality, clinical population, and acquisition setting. This enables evaluation of models' fairness under cross-domain shift, where performance and fairness must generalize beyond the training demographic distribution and modality. A representative sample from the dataset is shown in Fig.~\ref{fig:new-dataset-example}, and a detailed dataset analysis is presented in Fig.~\ref{fig:new-dataset-analysis}.

\textbf{Evaluation Metrics.} Same to~\cite{Fairclip-CVPR-2024}, we used the Area Under the Receiver Operating Characteristic Curve (AUC) as the overall performance metric, and the \textit{Demographic Parity Difference (DPD)}~\cite{RedFair-ICML-2018,FairReg-ICML-2019} and the \textit{Difference in Equalized Odds (DEOdds)}~\cite{RedFair-ICML-2018} for fairness evaluation. To highlight performance across different demographic groups, \textit{Group-wise AUC} is also reported. To capture the trade-off between performance and fairness, we also reported the \textit{Equity-Scaled AUC (ES-AUC)}~\cite{HarvardGF-TMI-2024}. Moreover, inspired by JTT~\cite{JTT-ICML-2021}, we reported the \textit{Worst-group AUC} to quantify the fairness to the worst-performing group. The formal mathematical definitions of these evaluation metrics are provided in \nameref{sec:appendix-a}. Finally, to complement the AUC-based analysis, we report a comprehensive set of additional results based on F1-score variants in \nameref{sec:appendix-b}. 

\subsection{Experimental Settings}
\label{subsec:exp-settings}

We conducted our experiments using CLIP~\cite{CLIP-ICML-2021}, employing the ViT-B/16 backbone \cite{ViT-ICLR-2021} for the visual encoder. The MLP in Eq. (\ref{eq:att-pred}) has three hidden layers of sizes 256, 128, and 64, respectively, each followed by ReLU activation. The fair pretraining process was conducted for 30 epochs using the Harvard-FairVLMed training set \cite{Fairclip-CVPR-2024} by an Adam optimizer with a weight decay of \(6 \times 10^{-5}\), betas of \(0.1\), and epsilon of \(10^{-6}\), utilizing a single NVIDIA Tesla V100 GPU. A grid search strategy was employed for hyperparameter tuning, with the optimal configuration summarized in Table~\ref{tab:hyperparameters}.

\begin{table}[h]
    \centering
    \caption{Hyperparameter Setting for FairEnc.}
    \label{tab:hyperparameters}
    \begin{tabular}{ll}
    \toprule
    \textbf{Hyperparameter} & \textbf{Value} \\
    \midrule
    Text-to-Text Contrastive lambda ($\lambda_{txt-txt}$) & 0.01 \\
    Adversarial lambda ($\lambda_{adv}$) & 1 \\
    Mutual Information lambda ($\lambda_{MI}$) & 1 \\
    Commitment lambda ($\lambda_{Cmt}$) & 0.25 \\
    Text-to-Recon-Image Contrastive lambda ($\lambda_{txt-\hat{img}}$) & 10 \\
    Neutral Note Primary Selection probability ($p_{txt_1}$) & 0.5 \\
    Dictionary size ($C$) & 64 \\
    Batch size ($B$) & 32 \\
    Learning rate & $10^{-5}$ \\
    Discriminators learning rate & $5 \times 10^{-5}$ \\
    \bottomrule
    \end{tabular}
\end{table}

\subsection{Experiments on Harvard-FairVLMed}
\textbf{Zero-short Inference.} We follow FairCLIP \cite{Fairclip-CVPR-2024}  and first report the zero-short inference performances on Harvard-FairVLMed \cite{Fairclip-CVPR-2024} achieved by ours and compare the performances with original CLIP \cite{CLIP-ICML-2021} and four state-of-the-art bias mitigation methods: Just Train Twice (JTT) \cite{JTT-ICML-2021}, FairCLIP \cite{Fairclip-CVPR-2024}, Robust FairCLIP~\cite{RobustFairCLIP-MIPR-2025}, and FairMoE~\cite{FairMoE-MICCAI-2025}. Table~\ref{tab:zero-shot-comparison-harvard} presents the results of zero-shot classification on the Harvard-FairVLMed test split. FairEnc outperformed all other methods in both overall performance and fairness metrics. It achieved the highest AUC, indicating superior classification performance, and the best Worst Group AUC across all sensitive attributes. Additionally, FairEnc demonstrated strong fairness, achieving the best DPD and DEOdds scores for all attributes except ethnicity. FairEnc also attained the highest ES-AUC scores across all four sensitive attributes, reflecting an optimal trade-off between performance and fairness. Furthermore, it achieved the best Group-wise AUC for every group within each sensitive attribute.

\textbf{Linear Probing.} We then use the visual features from the visual encoder of VLMs as inputs of a linear classifier, and update the parameters of the linear classifier for 1000 epochs with a learning rate of \(10^{-4}\) on Harvard-FairVLMed \cite{Fairclip-CVPR-2024} for glaucoma classification. The performances of FairEnc and other compared state-of-the-art methods are reported in  Table~\ref{tab:linear-probe-comparison-harvard}. FairEnc consistently demonstrates strong performance-fairness trade-offs across multiple sensitive attributes. In particular, FairEnc attains the highest AUC and ES‑AUC for Race and Ethnicity, while achieving competitive performance for Gender and Language. With respect to fairness metrics, FairEnc yields the lowest DPD for Race, Ethnicity, and Language, and the lowest DEOdds for Race and Language. Moreover, FairEnc achieves the best or among the best Worst‑Group AUC scores across most attributes and delivers consistently strong group‑wise AUC performance, with a limited number of exceptions for Gender and Language. Overall, these results highlight FairEnc’s ability to substantially improve fairness while maintaining competitive predictive performance across diverse demographic attributes.

\subsection{Experimental Results on FairFundus}

To evaluate the adaptability of FairEnc to a new imaging modality, we further conduct experiments on the FairFundus dataset. In this setting, we consider gender and age as sensitive attributes. For age, patients are categorized into three groups (0-30, 30-60, and 60-90), enabling an evaluation of fairness generalization to an unseen sensitive attribute that was not considered during fair pretraining. Owing to the substantial domain gap between the scanned laser ophthalmoscopy images in Harvard-FairVLMed~\cite{Fairclip-CVPR-2024} and the colorful retinal fundus images in FairFundus, zero-shot evaluation is infeasible; therefore, we adapt all foundation models to FairFundus using linear probing.

To accommodate the relatively small dataset size, we adopt a five-fold cross-validation strategy. Following standard preprocessing, optic disc regions are cropped from fundus images and converted to grayscale for consistency with Harvard-FairVLMed~\cite{Fairclip-CVPR-2024}. Linear probing is performed with a learning rate of \(5 \times 10^{-3}\). The results are reported in Table~\ref{tab:new-dataset-comparison}. Across all evaluated methods, FairEnc achieves the lowest Demographic Parity Difference (DPD) and Difference in Equalized Odds (DEOdds) for both gender and age, demonstrating strong fairness generalization across different image domains and populations. Notably, FairEnc maintains its fairness advantage for the age attribute despite age not being included during fair pretraining, highlighting its robustness to unseen sensitive attributes.

In contrast, several baseline methods achieve higher overall AUC, which also translates into improved ES-AUC, group-wise AUC, and worst‑group AUC scores. Despite this, FairEnc maintains diagnostic performance within a competitive range while delivering substantially improved fairness, reflecting the well‑known trade‑off between predictive performance and fairness. While FairEnc successfully mitigates this trade‑off on Harvard‑FairVLMed~\cite{Fairclip-CVPR-2024}, where both fairness and performance are improved simultaneously, the trade‑off re‑emerges under cross‑domain evaluation on FairFundus. Nevertheless, FairEnc consistently preserves superior fairness properties under domain shift, underscoring its robustness and suitability for equitable medical image analysis.

\begin{table*}[h]
    \centering
    \caption{Comparison between FairEnc and other State-of-the-art Methods using Zero-shot Testing on Harvard-FairVLMed. Means and standard deviations are reported over three trials.}
    \label{tab:zero-shot-comparison-harvard}
    \resizebox{\textwidth}{!}{
    \begin{tabular}{cccccccccccc}
    \toprule
    \textbf{Attribute} & \textbf{Model} & \textbf{DPD ↓} & \textbf{DEOdds ↓} & \textbf{AUC ↑} & \textbf{ES-AUC ↑} & \multicolumn{3}{c}{\textbf{Group-wise AUC ↑}} & \textbf{Worst Group AUC ↑} \\
    \midrule
    \multirow{7}{*}{\textbf{Race}} & & & & & & \textbf{Asian} & \textbf{Black} & \textbf{White} & \\
    & CLIP & 9.67  ±  4.95  &  12.99  ±  3.99  &  68.42  ±  0.88  &  62.58  ±  0.98  &  73.37  ±  1.82  &  70.17  ±  2.71  &  67.04  ±  1.25 &  67.04  ±  1.25 \\
    & JTT CLIP & 9.37  ±  2.25  &  16.48  ±  4.69  &  68.6  ±  2.83  &  64.1  ±  3.92  &  72.61  ±  3.36  &  70.38  ±  0.71  &  67.38  ±  3.47 &  67.38  ±  3.47 \\
    & FairCLIP & 5.45  ±  3.7  &  8.45  ±  5.61  &  66.47  ±  0.56  &  61.08  ±  0.38  &  70.4  ±  1.52  &  69.51  ±  1.5  &  64.6  ±  0.32 &  64.6  ±  0.32 \\
    & Robust FairCLIP & 8.99  ±  4.09  &  12.16  ±  5.42  &  66.06  ±  0.98  &  60.55  ±  0.39  &  70.77  ±  1.63  &  68.75  ±  3.38  &  64.62  ±  0.99  &  64.62  ±  0.99 \\
    & FairMoE & 11.97  ±  1.59  &  11.29  ±  1.69  &  67.15  ±  2.3  &  62.95  ±  2.48  &  70.35  ±  2.95  &  69.38  ±  1.85  &  65.86  ±  2.61  &  65.86  ±  2.61 \\
    & FairEnc (ours) & \textbf{3.2}  ±  0.89  &  \textbf{6.99}  ±  1.13  &  \textbf{73.98}  ±  3.0  &  \textbf{69.81}  ±  4.62  &  \textbf{76.65}  ±  2.2  &  \textbf{74.63}  ±  0.54  &  \textbf{73.37}  ±  4.02  &  \textbf{73.37}  ±  4.02 \\
    \midrule
    \multirow{7}{*}{\textbf{Gender}} & & & & & & \textbf{Female} & \textbf{Male} & & \\
    & CLIP & 3.48  ±  1.63  &  8.19  ±  2.39  &  68.42  ±  0.88  &  64.06  ±  1.57  &  65.42  ±  1.41  &  72.27  ±  0.63  & &  65.42  ±  1.41 \\
    & JTT CLIP & 3.87  ±  2.35  &  6.84  ±  3.43  &  68.6  ±  2.83  &  64.91  ±  2.57  &  66.09  ±  2.73  &  71.77  ±  2.91  & &  66.09  ±  2.73 \\
    & FairCLIP & 3.97  ±  1.67  &  7.71  ±  1.27  &  64.43  ±  5.32  &  61.09  ±  3.7  &  62.15  ±  4.26  &  67.47  ±  6.71  & &  62.15  ±  4.26 \\
    & Robust FairCLIP & 3.45  ±  1.46  &  6.56  ±  2.73  &  64.97  ±  0.76  &  61.56  ±  0.69  &  62.59  ±  0.68  &  68.13  ±  1.05  &   &  62.59  ±  0.68 \\
    & FairMoE & 2.32  ±  1.01  &  5.88  ±  1.64  &  72.71  ±  1.55  &  68.75  ±  1.17  &  70.15  ±  1.31  &  75.89  ±  1.8  &   &  70.15  ±  1.31 \\
    & FairEnc (ours) & \textbf{2.1}  ±  0.73  &  \textbf{5.29}  ±  2.32  &  \textbf{73.98}  ±  3.0  &  \textbf{69.65}  ±  3.1  &  \textbf{71.26}  ±  3.13  &  \textbf{77.49}  ±  2.79  &   &  \textbf{71.26}  ±  3.13 \\
    \midrule
    \multirow{7}{*}{\textbf{Ethnicity}} & & & & & & \textbf{Non-Hispanic} & \textbf{Hispanic} & & \\
    & CLIP & 5.27  ±  2.18  &  \textbf{6.94}  ±  5.36  &  68.42  ±  0.88  &  64.29  ±  1.73  &  68.65  ±  0.87  &  62.18  ±  2.47  & &  62.18  ±  2.47 \\
    & JTT CLIP & 6.08  ±  5.05  &  9.29  ±  7.61  &  68.6  ±  2.83  &  64.97  ±  2.49  &  68.78  ±  2.84  &  63.21  ±  2.53  & &  63.21  ±  2.53 \\
    & FairCLIP & \textbf{4.98}  ±  5.54  &  7.38  ±  7.92  &  63.48  ±  3.5  &  59.32  ±  2.21  &  63.73  ±  3.56  &  56.77  ±  1.85  & &  56.77  ±  1.85 \\
    & Robust FairCLIP & 9.41  ±  3.89  &  15.16  ±  6.48  &  67.26  ±  1.25  &  58.15  ±  2.17  &  67.85  ±  1.19  &  52.08  ±  3.6  &   &  52.08  ±  3.6 \\
    & FairMoE & 5.3  ±  2.28  &  7.08  ±  3.17  &  64.69  ±  7.83  &  61.73  ±  7.48  &  64.79  ±  7.94  &  61.66  ±  6.24  &   &  61.66  ±  6.24 \\
    & FairEnc (ours) & 7.31  ±  1.3  &  10.97  ±  4.06  &  \textbf{73.98}  ±  3.0  &  \textbf{67.86}  ±  4.24  &  \textbf{74.33}  ±  2.96  &  \textbf{65.12}  ±  5.45  &   &  \textbf{65.12}  ±  5.45 \\
    \midrule
    \multirow{7}{*}{\textbf{Language}} & & & & & & \textbf{English} & \textbf{Spanish} & \textbf{Others} & \\
    & CLIP & 10.1  ±  0.61  &  13.95  ±  0.3  &  68.42  ±  0.88  &  56.35  ±  1.79  &  68.69  ±  1.0  &  56.34  ±  4.23  &  59.17  ±  0.83 &  56.34  ±  4.23 \\
    & JTT CLIP & 12.79  ±  2.92  &  18.55  ±  5.08  &  68.6  ±  2.83  &  58.27  ±  3.96  &  68.81  ±  2.88  &  61.17  ±  8.04  &  58.91  ±  1.4 &  58.91  ±  1.4 \\
    & FairCLIP & 9.8  ±  4.34  &  20.09  ±  4.78  &  64.74  ±  2.21  &  54.58  ±  1.69  &  64.87  ±  2.28  &  53.13  ±  2.64  &  57.84  ±  1.51 &  53.13  ±  2.64 \\
    & Robust FairCLIP & 12.15  ±  1.14  &  19.94  ±  6.46  &  65.96  ±  1.88  &  52.73  ±  0.55  &  66.33  ±  1.72  &  49.24  ±  4.17  &  57.94  ±  4.58  &  49.24  ±  4.17 \\
    & FairMoE & 8.05  ±  3.29  &  15.38  ±  7.06  &  66.54  ±  8.89  &  56.9  ±  6.08  &  66.68  ±  9.43  &  63.92  ±  6.69  &  59.34  ±  0.88  &  59.34  ±  0.88 \\
    & FairEnc (ours) & \textbf{7.94}  ±  2.05  &  \textbf{11.61}  ±  2.19  &  \textbf{73.98}  ±  3.0  &  \textbf{61.56}  ±  3.09  &  \textbf{74.45}  ±  3.17  &  \textbf{67.99}  ±  8.53  &  \textbf{61.65}  ±  2.34  &  \textbf{61.65}  ±  2.34 \\
    \bottomrule
    \end{tabular}
    }
\end{table*}

\begin{table*}[h]
    \centering
    \caption{Comparison between FairEnc and other State-of-the-art Methods using Linear Probing Test on Harvard-FairVLMed. Means and standard deviations are reported over three trials.}
    \label{tab:linear-probe-comparison-harvard}
    \resizebox{\textwidth}{!}{
    \begin{tabular}{cccccccccccc}
    \toprule
    \textbf{Attribute} & \textbf{Model} & \textbf{DPD ↓} & \textbf{DEOdds ↓} & \textbf{AUC ↑} & \textbf{ES-AUC ↑} & \multicolumn{3}{c}{\textbf{Group-wise AUC ↑}} & \textbf{Worst Group AUC ↑} \\
    \midrule
    \multirow{7}{*}{\textbf{Race}} & & & & & & \textbf{Asian} & \textbf{Black} & \textbf{White} & \\
    & CLIP & 7.56  ±  4.08  &  13.96  ±  6.4  &  73.68  ±  1.95  &  70.46  ±  1.92  &  74.92  ±  2.8  &  73.28  ±  3.97  &  74.66  ±  1.54  &  73.28  ±  3.97 \\
    & JTT CLIP & 7.75  ±  3.41  &  10.2  ±  3.34  &  71.31  ±  1.54  &  68.51  ±  1.13  &  72.72  ±  1.11  &  71.13  ±  2.92  &  71.78  ±  2.04  &  71.13  ±  2.92 \\
    & FairCLIP & 10.15  ±  3.38  &  13.78  ±  3.46  &  72.94  ±  2.91  &  70.03  ±  2.81  &  74.2  ±  4.69  &  73.44  ±  2.14  &  73.57  ±  2.7  &  73.44  ±  2.14 \\
    & Robust FairCLIP & 5.67  ±  3.19  &  7.76  ±  2.69  &  73.69  ±  1.28  &  69.55  ±  2.73  &  76.96  ±  1.13  &  72.96  ±  3.21  &  73.99  ±  0.91  &  72.96  ±  3.21 \\
    & FairMoE & 10.82  ±  2.33  &  16.23  ±  3.62  &  74.75  ±  1.61  &  71.52  ±  0.83  &  76.84  ±  1.84  &  73.98  ±  0.71  &  75.85  ±  2.04  &  73.98  ±  0.71 \\
    & FairEnc (ours) & \textbf{5.49}  ±  2.57  &  \textbf{6.67}  ±  2.37  &  \textbf{75.64}  ±  1.09  &  \textbf{73.29}  ±  1.09  &  \textbf{77.54}  ±  1.2  &  \textbf{74.95}  ±  1.72  &  \textbf{75.89}  ±  1.28  &  \textbf{74.95}  ±  1.72 \\
    \midrule
    \multirow{7}{*}{\textbf{Gender}} & & & & & & \textbf{Female} & \textbf{Male} & & \\
    & CLIP & \textbf{1.35}  ±  0.5  &  6.28  ±  1.85  &  73.68  ±  1.95  &  70.04  ±  1.28  &  71.34  ±  1.56  &  76.52  ±  2.42  &   &  71.34  ±  1.56 \\
    & JTT CLIP & 1.76  ±  0.82  &  7.3  ±  1.33  &  71.31  ±  1.54  &  67.79  ±  1.76  &  68.97  ±  1.71  &  74.17  ±  1.37  &   &  68.97  ±  1.71 \\
    & FairCLIP & 2.76  ±  2.53  &  6.77  ±  1.7  &  73.79  ±  0.55  &  70.51  ±  0.91  &  71.67  ±  0.79  &  76.33  ±  0.25  &   &  71.67  ±  0.79 \\
    & Robust FairCLIP & 2.87  ±  1.63  &  7.27  ±  2.63  &  71.71  ±  2.17  &  68.24  ±  2.65  &  69.39  ±  2.58  &  74.49  ±  1.69  &   &  69.39  ±  2.58 \\
    & FairMoE & 2.11  ±  1.05  &  \textbf{6.17}  ±  1.29  &  \textbf{75.85}  ±  1.26  &  \textbf{72.14}  ±  0.55  &  \textbf{73.5}  ±  0.82  &  78.63  ±  1.78  &   &  \textbf{73.5}  ±  0.82 \\
    & FairEnc (ours) & 1.94  ±  0.44  &  7.41  ±  0.65  &  75.64  ±  1.09  &  71.65  ±  1.31  &  73.14  ±  1.29  &  \textbf{78.72}  ±  0.87  &   &  73.14  ±  1.29 \\
    \midrule
    \multirow{7}{*}{\textbf{Ethnicity}} & & & & & & \textbf{Non-Hispanic} & \textbf{Hispanic} & & \\
    & CLIP & 11.66  ±  0.37  &  15.8  ±  3.43  &  73.68  ±  1.95  &  67.22  ±  1.54  &  74.04  ±  1.97  &  64.43  ±  1.66  &   &  64.43  ±  1.66 \\
    & JTT CLIP & 11.85  ±  3.62  &  19.19  ±  3.01  &  71.31  ±  1.54  &  65.85  ±  3.17  &  71.64  ±  1.43  &  63.19  ±  4.41  &   &  63.19  ±  4.41 \\
    & FairCLIP & 10.29  ±  3.57  &  \textbf{14.2}  ±  6.22  &  74.02  ±  1.88  &  69.57  ±  2.53  &  74.27  ±  2.01  &  67.73  ±  3.88  &   &  67.73  ±  3.88 \\
    & Robust FairCLIP & 16.32  ±  3.33  &  23.04  ±  7.1  &  67.25  ±  2.54  &  61.43  ±  2.38  &  67.65  ±  2.72  &  57.95  ±  4.69  &   &  57.95  ±  4.69 \\
    & FairMoE & 13.52  ±  10.6  &  20.58  ±  11.93  &  67.67  ±  8.46  &  60.64  ±  4.64  &  68.09  ±  8.72  &  56.78  ±  4.6  &   &  56.78  ±  4.6 \\
    & FairEnc (ours) & \textbf{9.91}  ±  3.88  &  15.51  ±  4.38  &  \textbf{75.64}  ±  1.09  &  \textbf{69.97}  ±  1.54  &  \textbf{75.95}  ±  1.09  &  \textbf{67.82}  ±  1.98  &   &  \textbf{67.82}  ±  1.98 \\
    \midrule
    \multirow{7}{*}{\textbf{Language}} & & & & & & \textbf{English} & \textbf{Spanish} & \textbf{Others} & \\
    & CLIP & 7.13  ±  2.06  &  22.65  ±  9.14  &  73.68  ±  1.95  &  61.62  ±  1.92  &  74.43  ±  2.06  &  69.79  ±  4.16  &  61.06  ±  3.03  &  61.06  ±  3.03 \\
    & JTT CLIP & 9.12  ±  0.71  &  29.36  ±  5.95  &  71.31  ±  1.54  &  57.83  ±  7.16  &  71.98  ±  1.27  &  67.14  ±  15.29  &  58.27  ±  8.51  &  58.27  ±  8.51 \\
    & FairCLIP & 15.74  ±  5.31  &  24.1  ±  5.18  &  73.56  ±  1.3  &  \textbf{65.55}  ±  3.52  &  73.83  ±  1.56  &  73.01  ±  6.61  &  \textbf{67.41}  ±  2.66  &  \textbf{67.41}  ±  2.66 \\
    & Robust FairCLIP & 13.8  ±  3.64  &  26.99  ±  3.12  &  72.27  ±  3.32  &  61.9  ±  2.8  &  72.53  ±  3.75  &  \textbf{82.76}  ±  2.09  &  66.5  ±  3.85  &  66.5  ±  3.85 \\
    & FairMoE & 8.05  ±  3.69  &  27.13  ±  8.87  &  68.14  ±  8.14  &  56.39  ±  5.49  &  68.29  ±  8.72  &  76.23  ±  5.95  &  59.28  ±  2.98  &  59.28  ±  2.98 \\
    & FairEnc (ours) & \textbf{6.02}  ±  2.77  &  \textbf{18.6}  ±  5.68  &  \textbf{75.64}  ±  1.09  &  63.8  ±  1.85  &  \textbf{76.34}  ±  1.04  &  72.25  ±  1.28  &  61.11  ±  3.09  &  61.11  ±  3.09 \\
    \bottomrule
    \end{tabular}
    }
\end{table*}

\begin{table*}[h]
    \centering
    \caption{Comparison between FairEnc and other State-of-the-art Methods using 5 Fold Cross-Validation Linear Probing on FairFundus. Means and standard deviations are reported over three trials.}
    \label{tab:new-dataset-comparison}
    \resizebox{\textwidth}{!}{
    \begin{tabular}{cccccccccccc}
    \toprule
    \textbf{Attribute} & \textbf{Model} & \textbf{DPD ↓} & \textbf{DEOdds ↓} & \textbf{AUC ↑} & \textbf{ES-AUC ↑} & \multicolumn{3}{c}{\textbf{Group-wise AUC ↑}} & \textbf{Worst Group AUC ↑} \\
    \midrule
    \multirow{7}{*}{\textbf{Gender}} & & & & & & \textbf{Female} & \textbf{Male} & & \\
    & CLIP & 2.18  ±  0.48  &  3.31  ±  1.11  &  \textbf{75.14}  ±  0.69  &  70.44  ±  0.42  &  \textbf{76.01}  ±  0.39  &  72.88  ±  1.16  &   &  72.88  ±  1.16 \\
    & JTT CLIP & 2.22  ±  0.88  &  3.19  ±  1.08  &  75.02  ±  0.59  &  \textbf{71.81}  ±  0.46  &  75.06  ±  1.28  &  73.65  ±  0.77  &   &  73.65  ±  0.77 \\
    & FairCLIP & 2.55  ±  0.94  &  4.89  ±  1.99  &  75.08  ±  0.69  &  71.42  ±  0.9  &  73.78  ±  1.59  &  \textbf{74.63}  ±  0.36  &   &  \textbf{73.78}  ±  1.59 \\
    & Robust FairCLIP & 2.5  ±  0.29  &  4.58  ±  1.15  &  74.32  ±  0.36  &  70.25  ±  0.57  &  75.07  ±  1.7  &  72.45  ±  0.6  &   &  72.45  ±  0.6 \\
    & FairMoE & 1.14  ±  0.63  &  2.56  ±  1.32  &  71.58  ±  1.19  &  68.26  ±  1.18  &  72.14  ±  1.87  &  69.6  ±  1.05  &   &  69.6  ±  1.05 \\
    & FairEnc (ours) & \textbf{1.09}  ±  0.26  &  \textbf{2.53}  ±  0.19  &  71.66  ±  3.49  &  67.55  ±  3.65  &  71.82  ±  3.57  &  69.96  ±  3.2  &   &  69.96  ±  3.2 \\
    \midrule
    \multirow{7}{*}{\textbf{Age}} & & & & & & \textbf{0-30} & \textbf{30-60} & \textbf{60-90} & \\
    & CLIP & 4.15  ±  1.26  &  9.72  ±  1.29  &  \textbf{75.14}  ±  0.69  &  \textbf{65.33}  ±  0.67  &  \textbf{80.97}  ±  1.1  &  71.0  ±  1.07  &  71.79  ±  1.06  &  71.0  ±  1.07 \\
    & JTT CLIP & 2.83  ±  0.63  &  9.84  ±  3.16  &  75.02  ±  0.59  &  64.65  ±  1.05  &  76.78  ±  3.36  &  71.2  ±  0.65  &  \textbf{72.29}  ±  1.56  &  \textbf{71.2}  ±  0.65 \\
    & FairCLIP & 3.15  ±  0.62  &  14.62  ±  3.78  &  75.08  ±  0.69  &  64.72  ±  0.35  &  78.43  ±  1.04  &  72.7  ±  0.62  &  71.05  ±  0.12  &  71.05  ±  0.12 \\
    & Robust FairCLIP & 2.93  ±  0.44  &  10.19  ±  1.54  &  74.32  ±  0.36  &  64.97  ±  0.82  &  77.7  ±  1.79  &  \textbf{73.03}  ±  0.69  &  70.16  ±  0.74  &  70.16  ±  0.74 \\
    & FairMoE & 2.52  ±  0.92  &  8.06  ±  3.78  &  71.58  ±  1.19  &  61.64  ±  0.71  &  72.68  ±  5.19  &  69.86  ±  1.96  &  66.5  ±  1.51  &  66.5  ±  1.51 \\
    & FairEnc (ours) & \textbf{1.91}  ±  0.79  &  \textbf{6.5}  ±  1.1  &  71.66  ±  3.49  &  61.15  ±  5.31  &  73.48  ±  0.75  &  66.55  ±  5.4  &  68.92  ±  3.27  &  66.55  ±  5.4 \\
    \bottomrule
    \end{tabular}
    }
\end{table*}

\subsection{Ablation Studies}

We conducted comprehensive ablation studies to evaluate the contributions of each component in FairEnc, identify the most effective combinations of text description inputs, and determine the optimal values for key hyperparameters. For each ablation setting, we report three metrics: AUC (as an indicator of overall performance), and the average DPD and average DEOdds across the four sensitive attributes (as fairness metrics). All experiments were conducted using the same configuration described in Subsection~\ref{subsec:exp-settings} and each was repeated with three different random seeds, and the reported results are the mean and standard deviation of zero-shot testing on the Harvard-FairVLMed~\cite{Fairclip-CVPR-2024} test set.

\subsubsection{Ablation of FairEnc Components}

\textbf{Component Removal.}  
FairEnc integrates three key components: (1) contrastive loss between text feature pairs, (2) mutual information (MI) regularization, and (3) adversarial debiasing. We evaluated the impact of removing each component individually, as well as all possible pairs of components. Results in Table~\ref{tab:components-ablation} demonstrate that all components are essential and complementary. The best AUC was achieved when using only the text contrastive and adversarial components. However, this configuration significantly underperformed in fairness metrics compared to the full FairEnc model, which achieved the best average DPD and DEOdds, along with the second-best AUC highlighting the importance of incorporating all components.

\begin{table}[h]
    \centering
    \caption{Ablation Study for FairEnc Components.}
    \label{tab:components-ablation}
    \resizebox{\linewidth}{!}{
    \begin{tabular}{cccccc}
    \hline
    \textbf{Text Cont.} & \textbf{MI Reg.} & \textbf{Adversarial} & \textbf{Avg. DPD ↓} & \textbf{Avg. DEOdds ↓} & \textbf{AUC ↑} \\
    \hline
    \ding{51} &  &  & 9.91 ± 5.45 & 14.84 ± 7.62 & 70.68 ± 2.7 \\
     & \ding{51} &  & 9.44 ± 1.67 & 14.97 ± 5.54 & 73.1 ± 2.12 \\
     &  & \ding{51} & 8.58 ± 5.28 & 13.6 ± 4.74 & 70.3 ± 7.11 \\
    \ding{51} & \ding{51} &  & 6.39 ± 1.99 & 10.29 ± 2.78 & 73.92 ± 0.63 \\
    \ding{51} &  & \ding{51} & 7.02 ± 2.34 & 14.06 ± 2.69 & \textbf{74.97} ± 0.8 \\
     & \ding{51} & \ding{51} & 7.83 ± 1.94 & 12.7 ± 3.06 & 72.91 ± 0.88 \\
    \ding{51} & \ding{51} & \ding{51} & \textbf{5.14} ± 1.24 & \textbf{8.72} ± 2.42 & 73.98 ± 3.0 \\
    \hline
    \end{tabular}
    }
\end{table}

\textbf{Lambda Values.}  
We also explored different values for the weighting hyperparameters of the three loss terms: the text-to-text contrastive loss (\( \lambda_{txt-txt} \)), mutual information loss (\( \lambda_{MI} \)), and adversarial loss (\( \lambda_{adv} \)). For each, we tested the best value obtained during validation, as well as values obtained by multiplying and dividing it by 2 and 10. All other hyperparameters were fixed to the values listed in Table~\ref{tab:hyperparameters}. Results in Table~\ref{tab:lambdas-ablation} show that \( \lambda_{txt-txt} = 0.01 \) yielded the best overall performance and fairness. For both \( \lambda_{MI} \) and \( \lambda_{adv} \), a value of 1 achieved the best fairness (average DPD and DEOdds) and the second-best AUC, which was comparable to the best AUC obtained with less fair configurations. Based on these findings, the optimal lambda values used in FairEnc are \( \lambda_{txt-txt} = 0.01 \), \( \lambda_{MI} = 1 \), and \( \lambda_{adv} = 1 \).

\begin{table}[h]
    \centering
    \caption{Ablation Study of Loss Components Lambdas for FairEnc.}
    \label{tab:lambdas-ablation}
    \resizebox{\linewidth}{!}{
    \begin{tabular}{ccccc}
    \hline
    \textbf{Lambda} & \textbf{Value} & \textbf{Avg. DPD ↓} & \textbf{Avg. DEOdds ↓} & \textbf{AUC ↑} \\
    \hline
    \multirow{5}{*}{\textbf{Text Cont.}} & 0.001 &  10.58  ±  4.52  &  14.36  ±  4.12  &  61.71  ±  9.63 \\
    & 0.005 & 8.61  ±  3.0  &  13.74  ±  4.56  &  69.81  ±  2.55 \\
    & 0.01 & \textbf{5.14}  ±  1.24  &  \textbf{8.72}  ±  2.42  &  \textbf{73.98}  ±  3.0 \\
    & 0.02 & 7.72  ±  2.58  &  12.43  ±  2.55  &  69.06  ±  5.18 \\
    & 0.1 & 8.52  ±  4.39  &  12.56  ±  3.4  &  68.48  ±  2.62  \\
    \hline
    \multirow{5}{*}{\textbf{MI Reg.}} & 0.1 &  7.7  ±  2.76  &  12.64  ±  4.95  &  73.23  ±  3.11 \\
    & 0.5 & 10.68  ±  2.08  &  13.24  ±  3.89  &  70.38  ±  3.88 \\
    & 1 & \textbf{5.14}  ±  1.24  &  \textbf{8.72}  ±  2.42  &  73.98  ±  3.0 \\
    & 2 & 6.93  ±  2.86  &  12.66  ±  6.08  &  \textbf{75.24}  ±  0.67 \\
    & 10 & 10.27  ±  4.47  &  13.31  ±  4.7  &  67.98  ±  7.57  \\
    \hline
    \multirow{5}{*}{\textbf{Adversarial}} & 0.1 & 7.85  ±  2.49  &  13.93  ±  3.61  &  \textbf{75.04}  ±  0.83  \\
    & 0.5 & 9.09  ±  4.12  &  14.21  ±  4.67  &  68.63  ±  6.43 \\
    & 1 & \textbf{5.14}  ±  1.24  &  \textbf{8.72}  ±  2.42  &  73.98  ±  3.0 \\
    & 2 & 9.2  ±  2.64  &  15.96  ±  4.38  &  71.0  ±  0.6 \\
    & 10 & 9.74  ±  2.37  &  14.16  ±  4.16  &  70.56  ±  1.79  \\
    \hline
    \end{tabular}
    }
\end{table}

\subsubsection{Ablation of Text Descriptions}

\textbf{Text Description Types.}  
To assess the impact of different synthetic text description inputs, we evaluated all possible combinations of two text descriptions per sample. These included: (1) original + synthetic neutral, (2) original + synthetic with random demographics, (3) two synthetic versions with different random demographics, and (4) synthetic neutral + synthetic with random demographics. For each pair of different types, either description was randomly selected as the primary input $ \mathbf{X}_{txt_1}^{(i)} $ with 50\% probability, and the other as the secondary input $ \mathbf{X}_{txt_2}^{(i)} $. When using synthetic versions with random demographics, one or two versions were randomly selected from the five generated variants.

Results in Table~\ref{tab:text-types-ablation} show that pairs involving the original description consistently resulted in the worst fairness (average DPD and DEOdds), despite one of them achieving the best AUC. In contrast, pairs composed entirely of synthetic descriptions achieved better fairness. However, the pair using two random demographic versions had the lowest AUC, despite achieving the best average DEOdds. The best trade-off was achieved by the pair consisting of the synthetic neutral version and one random demographic version, which yielded the best average DPD and the second-best AUC and average DEOdds making it the most balanced and effective configuration.

\begin{table}[h]
    \centering
    \caption{Ablation Study of Text Types for FairEnc.}
    \label{tab:text-types-ablation}
    \resizebox{\linewidth}{!}{
    \begin{tabular}{cccccc}
    \hline
    \textbf{Original} & \textbf{Neutral} & \textbf{Random} & \textbf{Avg. DPD ↓} & \textbf{Avg. DEOdds ↓} & \textbf{AUC ↑} \\
    \hline
    \ding{51} & \ding{51} &  & 8.74  ±  4.68  &  13.33  ±  4.49  &  68.51  ±  8.87 \\
    \ding{51} &  & \ding{51} & 6.95  ±  2.96  &  12.6  ±  3.93  &  \textbf{75.24}  ±  0.08 \\
     &  & \ding{51} & 5.37  ±  5.59  &  \textbf{7.82}  ±  6.48  &  62.42  ±  8.2 \\
     & \ding{51} & \ding{51} & \textbf{5.14}  ±  1.24  &  8.72  ±  2.42  &  73.98  ±  3.0 \\
    \hline
    \end{tabular}
    }
\end{table}

\textbf{Random Selection Distributions.}  
The distributions described in Eq. (\ref{eq:text-select-dist}) controlling the random selection of the input synthesized text descriptions $ \mathbf{X}_{txt_1}^{(i)} $ and $ \mathbf{X}_{txt_2}^{(i)} $ depend on the neutral version primary selection probability $p_{txt_1}$. To determine the optimal distributions, we tested five values for $p_{txt_1}$ ranging from 0 to 1 in increments of 0.25. Results in Table~\ref{tab:neutral-prob-ablation} indicate that a probability of 0.5 achieved the best fairness (average DPD and DEOdds) and the second-best AUC, which was comparable to the best-performing configuration. This confirms that a balanced 50\% probability for selecting either the neutral or random demographic version as the primary input $ \mathbf{X}_{txt_1}^{(i)} $ yields the most effective trade-off between performance and fairness.

\begin{table}[h]
    \centering
    \caption{Ablation Study of Neutral Text Version Primary Selection Probability $p_{txt_1}$ for FairEnc.}
    \label{tab:neutral-prob-ablation}
    \resizebox{\linewidth}{!}{
    \begin{tabular}{cccc}
    \hline
    \textbf{Probability} & \textbf{Avg. DPD ↓} & \textbf{Avg. DEOdds ↓} & \textbf{AUC ↑} \\
    \hline
    0 &  8.17  ±  2.3  &  14.38  ±  5.36  &  \textbf{74.09}  ±  1.18 \\
    0.25 & 8.64  ±  6.1  &  11.86  ±  5.96  &  67.7  ±  8.28 \\
    0.5 & \textbf{5.14}  ±  1.24  &  \textbf{8.72}  ±  2.42  &  73.98  ±  3.0 \\
    0.75 & 7.54  ±  3.42  &  13.23  ±  3.74  &  71.79  ±  6.05 \\
    1 & 10.04  ±  4.16  &  14.68  ±  4.58  &  68.97  ±  4.16  \\
    \hline
    \end{tabular}
    }
\end{table}

\textbf{Synthetic Notes Generation LLM.}
The quality of the synthetic clinical notes plays a critical role in the learning of the fair text encoder. In particular, the generated notes must be well-formed, fluent, and minimally influenced by inherent biases present in the underlying generation model. These considerations guided the design of the prompts described in Subsection~\ref{subsec:fair-text}. To identify the most suitable LLM for synthetic notes generation, we compared three different LLMs: the state‑of‑the‑art general‑purpose models Llama~\cite{Llama-arXiv-2024} and Qwen~\cite{Qwen-arXiv-2025}, as well as the medically adapted LLM Med42~\cite{Med42-arXiv-2024}. For each model, we generated both the neutral and random demographic synthetic clinical notes following the procedure outlined in Subsection~\ref{subsec:fair-text}, and used the resulting notes for pretraining FairEnc.

The results reported in Table~\ref{tab:llm-ablation} indicate that Qwen~\cite{Qwen-arXiv-2025} consistently provides the best trade‑off between overall performance and fairness. In contrast, Med42~\cite{Med42-arXiv-2024} results in a noticeable degradation in overall performance measured by AUC, while Llama~\cite{Llama-arXiv-2024} leads to a clear deterioration in fairness assessed by average DPD and DEOdds. Based on these findings, we adopt Qwen‑generated synthetic clinical notes in our experiments.

\begin{table}[h]
    \centering
    \caption{Ablation Study of LLM used in Synthetic Notes Generation for FairEnc.}
    \label{tab:llm-ablation}
    \resizebox{\linewidth}{!}{
    \begin{tabular}{cccc}
    \hline
    \textbf{LLM} & \textbf{Avg. DPD ↓} & \textbf{Avg. DEOdds ↓} & \textbf{AUC ↑} \\
    \hline
    Med42\cite{Med42-arXiv-2024} & 5.53  ±  4.84  &  \textbf{7.48}  ±  5.98  &  66.49  ±  6.99 \\
    Llama\cite{Llama-arXiv-2024} & 7.23  ±  3.8  &  10.98  ±  1.86  &  71.28  ±  0.35 \\
    Qwen\cite{Qwen-arXiv-2025} & \textbf{5.14}  ±  1.24  &  8.72  ±  2.42  &  \textbf{73.98}  ±  3.0 \\
    \hline
    \end{tabular}
    }
\end{table}

\subsubsection{Ablation of Mutual Information Estimation}

Estimating mutual information (MI) becomes particularly challenging when one or more of the variables involved are high-dimensional and continuous, as is the case for the visual features $\mathbf{f}_{img}^{(i)}$ in our framework. In such settings, a number of neural-based approaches have been proposed to approximate mutual information, including Mutual Information Neural Estimation (MINE)~\cite{MINE-ICML-2018} as well as contrastive learning objectives such as InfoNCE~\cite{InfoNCE-arXiv-2018}. These methods provide tractable estimates of mutual information by maximizing a variational lower bound.

While lower-bound estimators are effective when mutual information is intended to be maximized, they are considerably less suitable in fairness-oriented objectives where mutual information must instead be minimized. In such cases, minimizing a lower bound does not guarantee any meaningful decrease in the true mutual information, as the bound can trivially collapse without imposing a strict restriction on the dependence between the learned representations and the sensitive attributes.

CLUB~\cite{CLUB-ICML-2020} addresses this limitation by providing a variational upper bound on mutual information through explicit modeling of the conditional distribution. Minimizing an upper bound yields a conservative but principled mechanism for suppressing information leakage, making CLUB~\cite{CLUB-ICML-2020} more aligned with fairness-driven learning objectives. Moreover, CLUB~\cite{CLUB-ICML-2020} relies on an auxiliary adversarial network to estimate the conditional likelihood, allowing it to be seamlessly integrated into our FairEnc framework by reusing the same discriminators employed for adversarial debiasing.

To evaluate the effectiveness of our proposed Dictionary Proxy-based Mutual Information estimator (Dict-Proxy), we conduct an ablation study comparing the use of CLUB~\cite{CLUB-ICML-2020} and Dict-Proxy as the mutual information estimator. As reported in Table~\ref{tab:mutual-information-ablation}, using Dict-Proxy consistently leads to improved fairness and performance. In particular, Dict-Proxy achieves better average DPD, average DEOdds, and AUC compared to CLUB-based estimation. These results demonstrate that our dictionary proxy provides a more effective and stable technique for minimizing sensitive information leakage in practice.

\begin{table}[h]
    \centering
    \caption{Ablation Study of Mutual Information Estimation Technique for FairEnc.}
    \label{tab:mutual-information-ablation}
    \resizebox{\linewidth}{!}{
    \begin{tabular}{cccc}
    \hline
    \textbf{MI Estimator} & \textbf{Avg. DPD ↓} & \textbf{Avg. DEOdds ↓} & \textbf{AUC ↑} \\
    \hline
    CLUB\cite{CLUB-ICML-2020} & 8.2  ±  4.0  &  11.83  ±  6.49  &  73.08  ±  1.22 \\
    Dict-Proxy & \textbf{5.14}  ±  1.24  &  \textbf{8.72}  ±  2.42  &  \textbf{73.98}  ±  3.0 \\
    \hline
    \end{tabular}
    }
\end{table}
\section{Discussion}

\subsection{Effect of Simultaneous Debiasing of Sensitive Attributes}

To study the interactions between sensitive attributes during simultaneous debiasing, we applied FairEnc to all combinations of the four attributes, using identical hyperparameters (Table~\ref{tab:hyperparameters}). Zero-shot classification results on the Harvard-FairVLMed test split are summarized in Table~\ref{tab:different-attributes-combinations}, demonstrating that our approach generally preserves strong performance and fairness across combinations.

Overall, jointly debiasing \textit{gender} and \textit{ethnicity} yields the most favorable outcomes. This combination achieves the best DPD and DEOdds for both attributes, along with the highest Ethnicity ES-AUC and Worst Group AUC, while remaining competitive on all other metrics. In contrast, simultaneously debiasing \textit{race} and \textit{language} leads to degraded performance and fairness, resulting in the lowest AUC, ES-AUC, and Worst Group AUC, as well as the worst DEOdds for language. Notably, debiasing gender and ethnicity together outperforms debiasing each independently, whereas the opposite trend is observed for race and language. Adding either gender or ethnicity to the race-language combination alleviates their negative interaction, with ethnicity providing a stronger corrective effect. Conversely, incorporating race or language into the gender-ethnicity combination weakens their positive synergy, particularly when race is included. When all four attributes are debiased jointly, the beneficial interaction between gender and ethnicity dominates, yielding a single VLM that is both highly performant and fair with respect to all attributes simultaneously.

\begin{figure}%[]
    \centering
    \includegraphics[width=\linewidth]{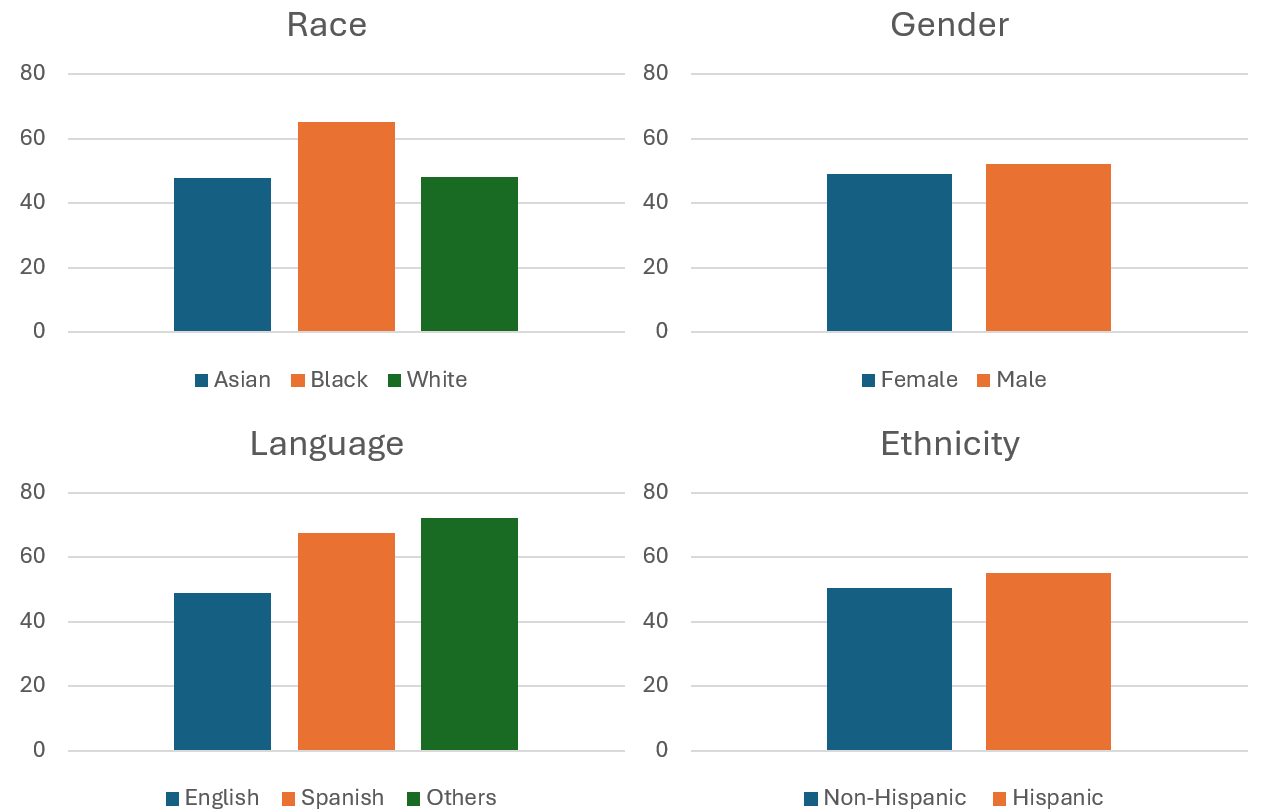}
    \caption{Glaucoma Relevance per Group of each Sensitive Attribute in Harvard-FairVLMed Training set. Relevance is the Proportion of Positive Samples within each Group.}
    \label{fig:groups-relevance}
\end{figure}

These behaviors are further explained by Fig.~\ref{fig:groups-relevance}, which shows that gender and ethnicity groups exhibit relatively balanced glaucoma relevance, whereas race and language groups display substantial imbalance. This imbalance likely causes interference between the adversarial and mutual information losses for race and language, potentially introducing interchangeable spurious correlations. In contrast, the balanced nature of gender and ethnicity facilitates cooperative optimization and enables a robust, fair VL model.

\begin{table*}[h]
    \centering
    \caption{Zero-shot Testing Results of FairEnc on Harvard-FairVLMed using Different Sensitive Attributes Combinations. The Debiased Attributes for each Model are listed in the 'Combination' column, with R representing Race, G representing Gender, E representing Ethnicity, and L representing Language, and Attributes are separated by spaces.}
    \label{tab:different-attributes-combinations}
    \resizebox{\textwidth}{!}{
    \begin{tabular}{cccccccccccc}
    \toprule
    \textbf{Attribute} & \textbf{Combination} & \textbf{DPD ↓} & \textbf{DEOdds ↓} & \textbf{AUC ↑} & \textbf{ES-AUC ↑} & \multicolumn{3}{c}{\textbf{Group-wise AUC ↑}} & \textbf{Worst Group AUC ↑} \\
    \midrule
    \multirow{9}{*}{\textbf{Race}} & & & & & & \textbf{Asian} & \textbf{Black} & \textbf{White} & \\
    & R & \textbf{2.89  ±  0.6} & 10.89  ±  2.5 & 73.3  ±  1.72 & 69.54  ±  0.28 & 76.28  ±  2.61 & 71.3  ±  0.69 & 73.57  ±  2.08 & 71.3  ±  0.69 \\
    & R G & 13.85  ±  10.29 & 14.55  ±  10.92 & 68.64  ±  8.29 & 66.31  ±  8.89 & 69.44  ±  7.97 & 68.27  ±  6.32 & 68.13  ±  9.44 & 68.13  ±  9.44 \\
    & R E & 6.8  ±  3.99 & 11.69  ±  2.7 & 71.33  ±  4.4 & 67.51  ±  4.71 & 74.65  ±  3.91 & 71.63  ±  3.07 & 70.99  ±  5.02 & 70.99  ±  5.02 \\
    & R L & 10.08  ±  13.46 & 13.63  ±  11.79 & 61.87  ±  9.97 & 58.9  ±  9.98 & 62.63  ±  11.49 & 63.15  ±  8.36 & 60.64  ±  11.18 & 60.64  ±  11.18 \\
    & R G E & 14.26  ±  8.52 & 13.69  ±  7.91 & 69.92  ±  7.47 & 66.05  ±  7.75 & 73.9  ±  7.37 & 69.61  ±  8.65 & 69.04  ±  7.89 & 69.04  ±  7.89 \\
    & R G L & 20.54  ±  9.0 & 20.84  ±  10.09 & 65.28  ±  8.06 & 62.23  ±  7.92 & 66.4  ±  6.4 & 65.14  ±  6.5 & 64.41  ±  9.32 & 64.41  ±  9.32 \\
    & R E L & 8.79  ±  4.04 & 9.59  ±  3.43 & 68.93  ±  3.63 & 66.33  ±  4.75 & 70.05  ±  2.99 & 67.15  ±  4.46 & 68.96  ±  4.01 & 67.15  ±  4.46 \\
    & R G E L & 3.2  ±  0.89 & \textbf{6.99  ±  1.13} & \textbf{73.98  ±  3.0} & \textbf{69.81  ±  4.62} & \textbf{76.65  ±  2.2} & \textbf{74.63  ±  0.54} & \textbf{73.37  ±  4.02} & \textbf{73.37  ±  4.02} \\
    \midrule
    \multirow{9}{*}{\textbf{Gender}} & & & & & & \textbf{Female} & \textbf{Male} & & \\
    & G & 2.97  ±  1.61 & 6.42  ±  1.92 & 69.52  ±  4.53 & 65.53  ±  4.13 & 66.85  ±  4.38 & 72.93  ±  4.67 & & 66.85  ±  4.38 \\
    & R G & 5.24  ±  2.68 & 9.75  ±  3.05 & 68.64  ±  8.29 & 65.64  ±  8.02 & 66.67  ±  8.27 & 71.26  ±  8.12 & & 66.67  ±  8.27 \\
    & G E & \textbf{1.06  ±  1.02} & \textbf{2.74  ±  2.45} & 73.63  ±  1.53 & 69.61  ±  1.59 & 71.04  ±  1.63 & 76.82  ±  1.43 & & 71.04  ±  1.63 \\
    & G L & 6.82  ±  2.27 & 10.92  ±  1.62 & 68.58  ±  6.58 & 65.12  ±  6.45 & 66.25  ±  6.63 & 71.59  ±  6.48 & & 66.25  ±  6.63 \\
    & R G E & 5.77  ±  1.21 & 10.88  ±  1.49 & 69.92  ±  7.47 & 66.06  ±  6.65 & 67.4  ±  7.13 & 73.18  ±  7.82 & & 67.4  ±  7.13 \\
    & R G L & 5.95  ±  2.5 & 9.75  ±  2.37 & 65.28  ±  8.06 & 61.92  ±  7.6 & 62.92  ±  7.95 & 68.34  ±  8.07 & & 62.92  ±  7.95 \\
    & G E L & 2.69  ±  1.79 & 6.11  ±  3.15 & 72.13  ±  2.11 & 68.44  ±  2.6 & 69.7  ±  2.54 & 75.12  ±  1.6 & & 69.7  ±  2.54 \\
    & R G E L & 2.1  ±  0.73 & 5.29  ±  2.32 & \textbf{73.98  ±  3.0} & \textbf{69.65  ±  3.1} & \textbf{71.26  ±  3.13} & \textbf{77.49  ±  2.79} & & \textbf{71.26  ±  3.13} \\
    \midrule
    \multirow{9}{*}{\textbf{Ethnicity}} & & & & & & \textbf{Non-Hispanic} & \textbf{Hispanic} & & \\
    & E & 9.45  ±  1.39 & 11.38  ±  1.26 & 71.46  ±  5.62 & 68.74  ±  6.1 & 71.59  ±  5.6 & 67.52  ±  6.77 & & 67.52  ±  6.77 \\
    & R E & 7.74  ±  2.79 & 12.3  ±  2.74 & 71.33  ±  4.4 & 67.98  ±  4.9 & 71.51  ±  4.35 & 66.49  ±  5.53 & & 66.49  ±  5.53 \\
    & G E & \textbf{5.58  ±  3.61} & \textbf{10.08  ±  5.0} & 73.63  ±  1.53 & \textbf{71.26  ±  3.6} & 73.76  ±  1.42 & \textbf{70.27  ±  4.61} & & \textbf{70.27  ±  4.61} \\
    & E L & 9.3  ±  3.48 & 13.77  ±  4.0 & \textbf{74.03  ±  2.97} & 69.1  ±  7.03 & 74.31  ±  2.72 & 66.43  ±  9.9 & & 66.43  ±  9.9 \\
    & R G E & 8.31  ±  5.42 & 13.04  ±  7.09 & 69.92  ±  7.47 & 67.39  ±  7.66 & 70.06  ±  7.46 & 66.23  ±  8.19 & & 66.23  ±  8.19 \\
    & R E L & 7.91  ±  5.67 & 11.6  ±  6.33 & 68.93  ±  3.63 & 65.49  ±  5.13 & 69.07  ±  3.48 & 64.82  ±  7.72 & & 64.82  ±  7.72 \\
    & G E L & 6.97  ±  2.92 & 10.79  ±  5.31 & 72.13  ±  2.11 & 67.91  ±  3.84 & 72.34  ±  2.0 & 66.69  ±  5.99 & & 66.69  ±  5.99 \\
    & R G E L & 7.31  ±  1.3 & 10.97  ±  4.06 & 73.98  ±  3.0 & 67.86  ±  4.24 & \textbf{74.33  ±  2.96} & 65.12  ±  5.45 & & 65.12  ±  5.45 \\
    \midrule
    \multirow{9}{*}{\textbf{Language}} & & & & & & \textbf{English} & \textbf{Spanish} & \textbf{Others} & \\
    & L & 12.39  ±  4.98 & 21.22  ±  6.6 & 70.18  ±  3.83 & 60.45  ±  1.32 & 70.43  ±  3.95 & 73.2  ±  8.46 & 59.11  ±  2.49 & 59.11  ±  2.49 \\
    & R L & 10.04  ±  7.51 & 24.71  ±  21.39 & 61.87  ±  9.97 & 53.15  ±  4.37 & 61.71  ±  10.47 & 69.98  ±  12.08 & 56.15  ±  0.78 & 56.15  ±  0.78 \\
    & G L & 14.23  ±  2.33 & 23.46  ±  6.24 & 68.58  ±  6.58 & 57.87  ±  0.51 & 68.97  ±  7.12 & 66.1  ±  7.17 & 57.39  ±  3.72 & 57.39  ±  3.72 \\
    & E L & 9.3  ±  3.48 & 13.77  ±  4.0 & \textbf{74.03  ±  2.97} & 59.68  ±  3.71 & \textbf{74.69  ±  3.1} & 67.14  ±  9.46 & 58.18  ±  0.64 & 58.18  ±  0.64 \\
    & R G L & 15.17  ±  4.82 & 14.25  ±  7.26 & 65.28  ±  8.06 & 57.43  ±  5.35 & 65.35  ±  8.69 & 65.15  ±  1.55 & 60.69  ±  2.11 & 60.69  ±  2.11 \\
    & R E L & 10.23  ±  4.1 & 14.19  ±  3.54 & 68.93  ±  3.63 & 61.39  ±  1.63 & 69.35  ±  3.74 & 65.81  ±  4.0 & 60.17  ±  0.36 & 60.17  ±  0.36 \\
    & G E L & 10.21  ±  2.39 & 21.15  ±  12.77 & 72.13  ±  2.11 & \textbf{62.57  ±  2.17} & 72.56  ±  2.53 & \textbf{74.53  ±  0.27} & 59.51  ±  6.65 & 59.51  ±  6.65 \\
    & R G E L & \textbf{7.94  ±  2.05} & \textbf{11.61  ±  2.19} & 73.98  ±  3.0 & 61.56  ±  3.09 & 74.45  ±  3.17 & 67.99  ±  8.53 & \textbf{61.65  ±  2.34} & \textbf{61.65  ±  2.34} \\
    \bottomrule
    \end{tabular}
    }
\end{table*}

\subsection{Limitations}

Despite its demonstrated effectiveness, this work has several limitations that merit discussion. First, although FairEnc exhibits strong fairness generalization, its superior predictive performance does not consistently generalize across domains. In particular, when evaluated on the FairFundus dataset, FairEnc does not achieve the highest overall AUC, which consequently impacts AUC-dependent metrics such as ES-AUC, group-wise AUC, and worst-group AUC. This observation reflects the re-emergence of the well-known performance-fairness trade-off under substantial domain shifts in imaging modality and data distribution.

Second, while FairEnc achieves the best fairness performance measured by DPD and DEOdds across three sensitive attributes under each evaluation protocol on the Harvard-FairVLMed dataset~\cite{Fairclip-CVPR-2024}, it does not uniformly attain the strongest fairness outcomes for all attributes. In particular, ethnicity remains the most challenging attribute under zero-shot evaluation, and gender exhibits comparatively reduced fairness under the linear probing setting, highlighting the inherent difficulty of achieving uniformly optimal fairness across different demographic factors.

Third, FairEnc incorporates multiple complementary debiasing mechanisms that must be jointly optimized. Although the resulting architecture remains compact, the interaction among several objectives introduces a nontrivial set of hyperparameters (Table~\ref{tab:hyperparameters}), necessitating careful tuning to balance predictive performance and fairness. Automating or simplifying this tuning process represents a promising direction for future work.

Finally, the FairFundus dataset presents limited demographic diversity with respect to attributes such as race, ethnicity, and preferred language, as it was collected from a single and relatively homogeneous population. While FairFundus enables evaluation of cross-modality generalization and fairness robustness for age and gender, it does not permit a comprehensive assessment of fairness generalization across more diverse demographic distributions. Future studies incorporating larger and more demographically diverse fundus datasets would allow for a more thorough evaluation of FairEnc’s generalization capabilities.
\section{Conclusion}

In summary, FairEnc delivers a VLM that is jointly debiased with respect to multiple demographic attributes, achieving strong fairness improvements while maintaining competitive diagnostic performance. The newly introduced \textit{FairFundus} dataset provides a valuable benchmark for evaluating fairness in VLMs for glaucoma classification, complementing Harvard‑FairVLMed~\cite{Fairclip-CVPR-2024} with a distinct imaging modality and population. Evaluation on FairFundus highlights FairEnc’s ability to preserve fairness advantages under domain shift, including generalization to unseen sensitive attributes. Moreover, our analysis of simultaneous debiasing indicates that maintaining balanced disease relevance across sensitive attribute groups in the training data is a key factor for enabling effective cooperative fair pretraining.

\printcredits

\section*{Ethics statement}

This work involved the use of the Harvard-FairVLMed and FairFundus datasets. The Harvard-FairVLMed dataset is publicly available and was used in accordance with its terms of use. it has been appropriately cited within the manuscript, and therefore no additional ethical approval was required. For the FairFundus dataset, the study was reviewed and approved by the Medical Research Ethics Committee, Institutional Review Board, Mansoura Faculty of Medicine, Mansoura University, Egypt. The approval was granted under approval code R.25.08.3315, dated 11 September 2025. Data collection and study procedures were conducted in accordance with the approved protocol and applicable ethical standards.
\section*{Acknowledgments}

This work was supported by the Finnish Doctoral Program Network in Artificial Intelligence, AI-DOC (decision number VN/3137/2024-OKM-6), and by the Research Council of Finland (formerly the Academy of Finland) under Research Fellow (grant number 355095), and by the University of Oulu and the Research Council of Finland, Profi6 336449. The authors also gratefully acknowledge the CSC-IT Center for Science, Finland, for providing the computational resources.

\appendix
\section*{Appendix A.}
\label{sec:appendix-a}

\setcounter{equation}{0}
\renewcommand{\theequation}{A.\arabic{equation}}

This appendix summarizes the mathematical formulations of the evaluation metrics used in the main manuscript. In addition to the fairness metrics reported in the primary results, we also define the Weighted F1 and Equity-Scaled F1 (ES-F1) scores, which are used for supplementary performance analysis.

\subsection*{Notation}

Let $y \in \{0,1\}$ denote the ground-truth glaucoma label, where $1$ indicates the presence of glaucoma and $0$ denotes the absence of disease. Let $\hat{y} \in \{0,1\}$ denote the predicted label obtained after thresholding the model output. Probabilities are estimated over the evaluation set and denoted by $p(\cdot)$. And as in main manuscript, $A_m$ denote the $m$-th sensitive attribute (e.g., race, gender, ethnicity, or language), and $\mathcal{A}_m$ denote the set of possible values of attribute $A_m$. A particular demographic group is denoted by $a \in \mathcal{A}_m$.

\subsection*{Demographic Parity Difference (DPD)}
\begin{equation}
\mathrm{DPD}_m
=
\max_{a \in \mathcal{A}_m}
p(\hat{y}=1 \mid A_m=a)
-
\min_{a \in \mathcal{A}_m}
p(\hat{y}=1 \mid A_m=a) \;,
\end{equation}
DPD measures disparities in positive prediction rates across demographic groups and reflects whether a model produces systematically different outputs for different populations, regardless of the true disease labels.

\subsection*{Difference in Equalized Odds (DEOdds)}
\begin{equation}
\mathrm{DEOdds}_m
=
\max
\left(
\begin{aligned}
&\max_{a \in \mathcal{A}_m} \mathrm{TPR}_{m,a}
-
\min_{a \in \mathcal{A}_m} \mathrm{TPR}_{m,a}, \\
&\max_{a \in \mathcal{A}_m} \mathrm{FPR}_{m,a}
-
\min_{a \in \mathcal{A}_m} \mathrm{FPR}_{m,a}
\end{aligned}
\right) \;,
\end{equation}

where
\begin{equation}
\begin{aligned}
\mathrm{TPR}_{m,a} &= p(\hat{y}=1 \mid y=1, A_m=a) \;, \\
\mathrm{FPR}_{m,a} &= p(\hat{y}=1 \mid y=0, A_m=a) \;,
\end{aligned}
\end{equation}
DEOdds jointly evaluates disparities in sensitivity and specificity across demographic groups and is particularly relevant in medical diagnosis, where unequal error rates may lead to inequitable clinical outcomes.

\subsection*{Equity-Scaled AUC (ES-AUC)}
\begin{equation}
\mathrm{ES\text{-}AUC}_m
=
\frac{
\mathrm{AUC}
}{
1 + \sum\limits_{a \in \mathcal{A}_m}
\left|
\mathrm{AUC}
-
\mathrm{AUC}_{m,a}
\right|
} \;,
\end{equation}
Here, $\mathrm{AUC}$ denotes the overall area under the ROC curve computed on the full evaluation set, and $\mathrm{AUC}_{m,a}$ denotes the subgroup-specific AUC. ES-AUC penalizes performance imbalance across demographic groups, encouraging consistent diagnostic accuracy and representing a trade-off metric between performance and fairness.

\subsection*{Weighted F1 Score}
\begin{equation}
\mathrm{F1}
=
\sum_{c}
w_c
\frac{2\,\mathrm{TP}_c}
{2\,\mathrm{TP}_c + \mathrm{FP}_c + \mathrm{FN}_c},
\qquad
w_c = \frac{|\mathcal{D}_c|}{|\mathcal{D}|} \;,
\end{equation}
Here, $c$ indexes class labels corresponding to the positive and negative glaucoma classes. $\mathrm{TP}_c$, $\mathrm{FP}_c$, and $\mathrm{FN}_c$ denote the numbers of true positives, false positives, and false negatives for class $c$, respectively. $|\mathcal{D}_c|$ represents the number of samples belonging to class $c$, and $|\mathcal{D}|$ denotes the total number of samples in the dataset. So, the weighted F1 summarizes overall classification performance while accounting for class imbalance.

\subsection*{Equity-Scaled F1 (ES-F1)}
\begin{equation}
\mathrm{ES\text{-}F1}_m
=
\frac{
\mathrm{F1}
}{
1 + \sum\limits_{a \in \mathcal{A}_m}
\left|
\mathrm{F1}
-
\mathrm{F1}_{m,a}
\right|
} \;,
\end{equation}
Here, $\mathrm{F1}_{m,a}$ denotes the F1 score computed for subgroup $a$ of sensitive attribute $A_m$. ES-F1 penalizes disparities in F1 performance across demographic groups providing another trade-off metric between performance and fairness.
\section*{Appendix B.}
\label{sec:appendix-b}

\setcounter{table}{0}
\renewcommand{\thetable}{B.\arabic{table}}

This appendix provides additional comparisons between FairEnc and other state-of-the-art techniques employing Weighted F1-based metrics, complementing the primary results reported using AUC. Specifically, we present the overall F1, Equity-Scaled F1 (ES-F1), Group-wise F1, and Worst-group F1 scores across all evaluation protocols considered in the main manuscript. These include both zero-shot and linear probing evaluations on the Harvard-FairVLMed~\cite{Fairclip-CVPR-2024} dataset, as well as five-fold cross-validation linear probing experiments on the FairFundus dataset. The results reported in Tables~\ref{tab:f1-zero-shot-comparison-harvard}, \ref{tab:f1-linear-probe-comparison-harvard}, and \ref{tab:f1-new-dataset-comparison} confirm the conclusions drawn from the AUC-based analysis. In particular, FairEnc consistently yields a single VL model that outperforms competing methods on the Harvard-FairVLMed~\cite{Fairclip-CVPR-2024} dataset under both evaluation settings. While this superior overall performance does not fully generalize to the FairFundus dataset, FairEnc maintains its fairness advantage, as evidenced by the DPD and DEOdds metrics reported in the main results.

\begin{table*}[h]
    \centering
    \caption{Comparison between FairEnc and other State-of-the-art Methods using Zero-shot Testing on Harvard-FairVLMed based on F1 Scores. Means and standard deviations are reported over three trials.}
    \label{tab:f1-zero-shot-comparison-harvard}
    \resizebox{\textwidth}{!}{
    \begin{tabular}{cccccccccc}
    \toprule
    \textbf{Attribute} & \textbf{Model} & \textbf{F1 ↑} & \textbf{ES-F1 ↑} & \multicolumn{3}{c}{\textbf{Group-wise F1 ↑}} & \textbf{Worst Group F1 ↑} \\
    \midrule
    \multirow{7}{*}{\textbf{Race}} & & & & \textbf{Asian} & \textbf{Black} & \textbf{White} & \\
    & CLIP & 63.99  ±  0.83  &  60.58  ±  0.59  &  67.41  ±  1.17  &  65.09  ±  2.13  &  63.39  ±  0.69  &  63.39  ±  0.69 \\
    & JTT CLIP & 64.45  ±  1.9  &  62.38  ±  2.04  &  65.17  ±  2.77  &  65.89  ±  1.55  &  64.1  ±  1.98  &  64.1  ±  1.98 \\
    & FairCLIP & 63.45  ±  0.25  &  60.25  ±  0.75  &  66.13  ±  1.67  &  65.06  ±  2.4  &  62.82  ±  0.16  &  62.82  ±  0.16 \\
    & Robust FairCLIP & 63.08  ±  0.87  &  59.94  ±  0.5  &  65.65  ±  3.11  &  63.32  ±  1.53  &  62.66  ±  1.06  &  62.66  ±  1.06 \\
    & FairMoE & 63.21  ±  1.33  &  60.24  ±  2.09  &  65.83  ±  2.85  &  63.61  ±  0.33  &  62.91  ±  1.69  &  62.91  ±  1.69 \\
    & FairEnc (ours) & \textbf{67.75}  ±  2.0  &  \textbf{65.05}  ±  3.42  &  \textbf{67.58}  ±  3.34  &  \textbf{69.76}  ±  0.54  &  \textbf{67.38}  ±  2.5  &  \textbf{67.38}  ±  2.5 \\
    \midrule
    \multirow{7}{*}{\textbf{Gender}} & & & & \textbf{Female} & \textbf{Male} & & \\
    & CLIP & 63.99  ±  0.83  &  59.8  ±  0.42  &  60.81  ±  0.46  &  67.82  ±  1.54  &   &  60.81  ±  0.46 \\
    & JTT CLIP & 64.45  ±  1.9  &  62.25  ±  1.42  &  62.84  ±  1.6  &  66.35  ±  2.29  &   &  62.84  ±  1.6 \\
    & FairCLIP & 62.52  ±  4.37  &  59.72  ±  3.32  &  60.41  ±  3.69  &  65.0  ±  5.24  &   &  60.41  ±  3.69 \\
    & Robust FairCLIP & 62.48  ±  0.68  &  59.33  ±  0.77  &  60.05  ±  0.78  &  65.37  ±  0.66  &   &  60.05  ±  0.78 \\
    & FairMoE & 67.19  ±  0.64  &  64.1  ±  0.56  &  65.01  ±  0.61  &  69.84  ±  0.7  &   &  65.01  ±  0.61 \\
    & FairEnc (ours) & \textbf{67.75}  ±  2.0  &  \textbf{64.51}  ±  2.12  &  \textbf{65.46}  ±  2.14  &  \textbf{70.51}  ±  2.08  &   &  \textbf{65.46}  ±  2.14 \\
    \midrule
    \multirow{7}{*}{\textbf{Ethnicity}} & & & & \textbf{Non-Hispanic} & \textbf{Hispanic} & & \\
    & CLIP & 63.99  ±  0.83  &  60.65  ±  1.48  &  64.2  ±  0.8  &  58.65  ±  2.01  &   &  58.65  ±  2.01 \\
    & JTT CLIP & 64.45  ±  1.9  &  61.82  ±  2.16  &  64.6  ±  1.88  &  60.32  ±  2.44  &   &  60.32  ±  2.44 \\
    & FairCLIP & 61.31  ±  3.19  &  58.91  ±  2.87  &  61.46  ±  3.19  &  57.4  ±  2.93  &   &  57.4  ±  2.93 \\
    & Robust FairCLIP & 63.76  ±  0.76  &  57.81  ±  2.69  &  64.16  ±  0.64  &  53.69  ±  4.57  &   &  53.69  ±  4.57 \\
    & FairMoE & 60.84  ±  5.66  &  58.81  ±  6.1  &  60.83  ±  5.77  &  60.35  ±  3.28  &   &  60.35  ±  3.28 \\
    & FairEnc (ours) & \textbf{67.75}  ±  2.0  &  \textbf{64.33}  ±  3.9  &  \textbf{67.96}  ±  1.94  &  \textbf{62.42}  ±  5.26  &   &  \textbf{62.42}  ±  5.26 \\
    \midrule
    \multirow{7}{*}{\textbf{Language}} & & & & \textbf{English} & \textbf{Spanish} & \textbf{Others} & \\
    & CLIP & 63.99  ±  0.83  &  56.76  ±  2.5  &  64.51  ±  0.79  &  56.96  ±  3.11  &  58.63  ±  2.19  &  56.96  ±  3.11 \\
    & JTT CLIP & 64.45  ±  1.9  &  58.21  ±  3.03  &  64.87  ±  1.92  &  62.09  ±  5.07  &  59.32  ±  3.98  &  59.32  ±  3.98 \\
    & FairCLIP & 62.94  ±  1.26  &  56.76  ±  1.57  &  63.32  ±  1.36  &  54.93  ±  3.97  &  60.72  ±  2.65  &  54.93  ±  3.97 \\
    & Robust FairCLIP & 62.54  ±  1.95  &  53.86  ±  1.06  &  63.12  ±  1.89  &  52.76  ±  2.43  &  56.82  ±  4.39  &  52.76  ±  2.43 \\
    & FairMoE & 62.26  ±  6.25  &  56.56  ±  7.03  &  62.46  ±  6.64  &  62.05  ±  2.39  &  60.82  ±  3.05  &  60.82  ±  3.05 \\
    & FairEnc (ours) & \textbf{67.75}  ±  2.0  &  \textbf{62.63}  ±  4.44  &  \textbf{68.16}  ±  1.84  &  \textbf{66.12}  ±  5.5  &  \textbf{62.49}  ±  4.21  &  \textbf{62.49}  ±  4.21 \\
    \bottomrule
    \end{tabular}
    }
\end{table*}

\begin{table*}[h]
    \centering
    \caption{Comparison between FairEnc and other State-of-the-art Methods using Linear Probing Test on Harvard-FairVLMed based on F1 Scores. Means and standard deviations are reported over three trials.}
    \label{tab:f1-linear-probe-comparison-harvard}
    \resizebox{\textwidth}{!}{
    \begin{tabular}{cccccccccc}
    \toprule
    \textbf{Attribute} & \textbf{Model} & \textbf{F1 ↑} & \textbf{ES-F1 ↑} & \multicolumn{3}{c}{\textbf{Group-wise F1 ↑}} & \textbf{Worst Group F1 ↑} \\
    \midrule
    \multirow{7}{*}{\textbf{Race}} & & & & \textbf{Asian} & \textbf{Black} & \textbf{White} & \\
    & CLIP & 67.82  ±  1.62  &  64.75  ±  1.36  &  70.3  ±  2.96  &  66.24  ±  2.23  &  67.98  ±  1.62  &  66.24  ±  2.23 \\
    & JTT CLIP & 65.3  ±  1.14  &  63.28  ±  0.64  &  66.34  ±  2.15  &  66.04  ±  1.01  &  65.17  ±  1.31  &  65.17  ±  1.31 \\
    & FairCLIP & 67.04  ±  2.16  &  63.4  ±  1.85  &  70.15  ±  5.07  &  65.47  ±  2.02  &  67.11  ±  2.06  &  65.47  ±  2.02 \\
    & Robust FairCLIP & 68.39  ±  0.69  &  64.97  ±  1.72  &  70.72  ±  2.73  &  66.54  ±  2.15  &  68.64  ±  0.54  &  66.54  ±  2.15 \\
    & FairMoE & 68.75  ±  1.04  &  65.07  ±  0.53  &  70.21  ±  1.03  &  65.16  ±  0.86  &  \textbf{69.38}  ±  1.38  &  65.16  ±  0.86 \\
    & FairEnc (ours) & \textbf{69.04}  ±  1.04  &  \textbf{66.37}  ±  0.66  &  \textbf{71.23}  ±  1.1  &  \textbf{68.49}  ±  0.8  &  69.04  ±  1.43  &  \textbf{68.49}  ±  0.8 \\
    \midrule
    \multirow{7}{*}{\textbf{Gender}} & & & & \textbf{Female} & \textbf{Male} & & \\
    & CLIP & 67.82  ±  1.62  &  64.94  ±  0.73  &  65.82  ±  0.99  &  70.25  ±  2.48  &   &  65.82  ±  0.99 \\
    & JTT CLIP & 65.3  ±  1.14  &  62.12  ±  0.8  &  62.99  ±  0.88  &  68.12  ±  1.62  &   &  62.99  ±  0.88 \\
    & FairCLIP & 68.06  ±  0.3  &  65.06  ±  0.25  &  65.97  ±  0.28  &  70.59  ±  0.34  &   &  65.97  ±  0.28 \\
    & Robust FairCLIP & 66.29  ±  1.18  &  63.73  ±  1.9  &  64.45  ±  1.76  &  68.51  ±  0.51  &   &  64.45  ±  1.76 \\
    & FairMoE & 68.95  ±  0.66  &  \textbf{66.3}  ±  0.3  &  \textbf{67.15}  ±  0.36  &  71.15  ±  1.1  &   &  \textbf{67.15}  ±  0.36 \\
    & FairEnc (ours) & \textbf{69.04}  ±  1.04  &  65.68  ±  1.44  &  66.72  ±  1.36  &  \textbf{71.86}  ±  0.64  &   &  66.72  ±  1.36 \\
    \midrule
    \multirow{7}{*}{\textbf{Ethnicity}} & & & & \textbf{Non-Hispanic} & \textbf{Hispanic} & & \\
    & CLIP & 67.82  ±  1.62  &  64.39  ±  1.52  &  68.02  ±  1.69  &  62.66  ±  2.21  &   &  62.66  ±  2.21 \\
    & JTT CLIP & 65.3  ±  1.14  &  60.84  ±  2.73  &  65.58  ±  1.06  &  58.1  ±  4.16  &   &  58.1  ±  4.16 \\
    & FairCLIP & 68.07  ±  1.29  &  65.14  ±  1.67  &  68.17  ±  1.36  &  65.07  ±  4.44  &   &  65.07  ±  4.44 \\
    & Robust FairCLIP & 63.33  ±  2.3  &  57.64  ±  1.71  &  63.68  ±  2.49  &  53.68  ±  3.8  &   &  53.68  ±  3.8 \\
    & FairMoE & 63.56  ±  5.82  &  58.51  ±  2.4  &  63.85  ±  6.02  &  55.44  ±  0.74  &   &  55.44  ±  0.74 \\
    & FairEnc (ours) & \textbf{69.04}  ±  1.04  &  \textbf{66.85}  ±  0.49  &  \textbf{69.17}  ±  1.09  &  \textbf{65.88}  ±  0.7  &   &  \textbf{65.88}  ±  0.7 \\
    \midrule
    \multirow{7}{*}{\textbf{Language}} & & & & \textbf{English} & \textbf{Spanish} & \textbf{Others} & \\
    & CLIP & 67.82  ±  1.62  &  61.62  ±  4.15  &  68.08  ±  1.78  &  71.68  ±  6.72  &  63.7  ±  0.99  &  63.7  ±  0.99 \\
    & JTT CLIP & 65.3  ±  1.14  &  58.13  ±  2.39  &  65.67  ±  0.95  &  65.4  ±  9.31  &  60.69  ±  3.56  &  60.69  ±  3.56 \\
    & FairCLIP & 67.86  ±  1.15  &  \textbf{64.42}  ±  1.37  &  67.88  ±  1.17  &  67.44  ±  2.98  &  \textbf{69.5}  ±  1.81  &  \textbf{67.44}  ±  2.98 \\
    & Robust FairCLIP & 67.01  ±  2.67  &  59.87  ±  3.98  &  67.11  ±  2.88  &  \textbf{73.73}  ±  6.02  &  65.21  ±  1.43  &  65.21  ±  1.43 \\
    & FairMoE & 63.06  ±  5.92  &  54.79  ±  5.59  &  63.17  ±  6.25  &  71.55  ±  5.11  &  60.38  ±  2.12  &  60.38  ±  2.12 \\
    & FairEnc (ours) & \textbf{69.04}  ±  1.04  &  62.31  ±  3.69  &  \textbf{69.55}  ±  0.95  &  71.3  ±  0.55  &  60.67  ±  4.99  &  60.67  ±  4.99 \\
    \bottomrule
    \end{tabular}
    }
\end{table*}

\begin{table*}[h]
    \centering
    \caption{Comparison between FairEnc and other State-of-the-art Methods using 5 Fold Cross-Validation Linear Probing on FairFundus based on F1 Scores. Means and standard deviations are reported over three trials.}
    \label{tab:f1-new-dataset-comparison}
    \resizebox{\textwidth}{!}{
    \begin{tabular}{cccccccccc}
    \toprule
    \textbf{Attribute} & \textbf{Model} & \textbf{F1 ↑} & \textbf{ES-F1 ↑} & \multicolumn{3}{c}{\textbf{Group-wise F1 ↑}} & \textbf{Worst Group F1 ↑} \\
    \midrule
    \multirow{7}{*}{\textbf{Gender}} & & & & \textbf{Female} & \textbf{Male} & & \\
    & CLIP & 70.15  ±  0.16  &  66.34  ±  0.18  &  73.37  ±  0.04  &  67.57  ±  0.14  &   &  67.57  ±  0.14 \\
    & JTT CLIP & 71.16  ±  0.47  &  67.36  ±  0.54  &  \textbf{73.55}  ±  0.93  &  69.2  ±  1.04  &   &  69.2  ±  1.04 \\
    & FairCLIP & \textbf{71.17}  ±  0.64  &  \textbf{67.54}  ±  0.83  &  72.09  ±  0.94  &  \textbf{70.21}  ±  0.59  &   &  \textbf{70.21}  ±  0.59 \\
    & Robust FairCLIP & 69.86  ±  0.24  &  65.04  ±  0.53  &  72.9  ±  1.08  &  67.24  ±  0.62  &   &  67.24  ±  0.62 \\
    & FairMoE & 68.97  ±  1.29  &  65.24  ±  1.55  &  71.98  ±  0.86  &  66.57  ±  1.6  &   &  66.57  ±  1.6 \\
    & FairEnc (ours) & 68.52  ±  2.25  &  64.34  ±  2.67  &  71.99  ±  1.6  &  65.78  ±  2.79  &   &  65.78  ±  2.79 \\
    \midrule
    \multirow{7}{*}{\textbf{Age}} & & & & \textbf{0-30} & \textbf{30-60} & \textbf{60-90} & \\
    & CLIP & 70.15  ±  0.16  &  57.3  ±  1.0  &  \textbf{84.28}  ±  1.28  &  67.94  ±  0.96  &  64.81  ±  1.22  &  64.81  ±  1.22 \\
    & JTT CLIP & 71.16  ±  0.47  &  59.19  ±  0.31  &  83.71  ±  0.32  &  69.28  ±  0.32  &  \textbf{66.8}  ±  1.24  &  \textbf{66.8}  ±  1.24 \\
    & FairCLIP & \textbf{71.17}  ±  0.64  &  \textbf{59.32}  ±  1.7  &  83.55  ±  0.94  &  \textbf{70.7}  ±  0.95  &  64.89  ±  1.69  &  64.89  ±  1.69 \\
    & Robust FairCLIP & 69.86  ±  0.24  &  57.14  ±  0.61  &  82.46  ±  1.08  &  69.66  ±  0.71  &  63.51  ±  0.73  &  63.51  ±  0.73 \\
    & FairMoE & 68.97  ±  1.29  &  55.94  ±  1.82  &  82.84  ±  0.85  &  68.69  ±  2.16  &  61.66  ±  1.5  &  61.66  ±  1.5 \\
    & FairEnc (ours) & 68.52  ±  2.25  &  56.03  ±  4.24  &  82.77  ±  1.28  &  66.05  ±  1.96  &  63.73  ±  4.27  &  63.73  ±  4.27 \\
    \bottomrule
    \end{tabular}
    }
\end{table*}

\clearpage

\bibliographystyle{cas-model2-names}

\bibliography{cas-refs}



\end{document}